\documentclass{article}

\usepackage{amsmath}
\usepackage{amsfonts}
\usepackage{amsthm}
\usepackage{microtype}
\usepackage{marginnote}
\usepackage{xcolor}
\usepackage{subcaption}
\usepackage{tikz}
\usepackage{dialogue}
\usepackage{appendix}
\usepackage{booktabs}
\usepackage{listings}
\usepackage{graphics}
\usepackage[framemethod=TikZ]{mdframed}
\usepackage[hidelinks]{hyperref}
\usepackage[letterpaper,margin=1.75in]{geometry}
\usepackage[font={small,it}]{caption}

\graphicspath{{./figures/}}

\let\citep\cite
\let\citet\cite

\usepackage{expl3}[2012-07-08]
\ExplSyntaxOn
\cs_new_eq:NN \fpeval \fp_eval:n
\ExplSyntaxOff

\DeclareMathOperator*{\argmin}{arg\,min}
\DeclareMathOperator{\codim} {codim}

\newtheorem{example}{Example}
\newtheorem{theorem}[example]{Theorem}

\newtheorem{proposition}[example]{Proposition}

\newtheorem{definition}[example]{Definition}
\newtheorem{assu}[example]{Assumption}

\newcommand{\pc}{\%}

\newcommand{\Etr}{\mathcal{E}_\text{tr}}
\newcommand{\R}{\mathbb{R}}

\newcommand\RR{\mathbb{R}}
\newcommand\SSS{\mathbb{S}}
\newcommand\matS{\mathcal{S}}
\newcommand\matC{\mathcal{C}}
\newcommand\NN{\mathbb{N}}

\newcommand\EE{\mathbb{E}}

\newcommand\Eall{{\mathcal{E}_\text{all}}}
\newcommand\Etrain{{\mathcal{E}_\text{tr}}}
\newcommand\PA{\mathrm{Pa}}

\usetikzlibrary{decorations.pathmorphing,shapes}

\newcommand{\ce}{Eric}
\newcommand{\ci}{Irma}
\newcommand{\se}{\speak{\ce}}
\newcommand{\si}{\speak{\ci}}
\newcommand{\dialogueseparator}{\begin{center} $\sim$ \end{center}}

\title{Invariant Risk Minimization}
\author{Martin Arjovsky, L\'eon Bottou, Ishaan Gulrajani, David Lopez-Paz}
\date{}

\begin{document}

\maketitle

\section{Introduction}

Machine learning suffers from a fundamental problem.
While machines are able to learn complex prediction rules by minimizing their training error, data are often marred by selection biases, confounding factors, and other peculiarities \citep{torralba2011unbiased, sturm2014simple, jabri2016revisiting}.
As such, machines justifiably inherit these data biases.
This limitation plays an essential role in the situations where machine learning fails to fulfill the promises of artificial intelligence.
More specifically, minimizing training error leads machines into recklessly absorbing all the correlations found in training data.
Understanding which patterns are useful has been previously studied as a correlation-versus-causation dilemma, since spurious correlations stemming from data biases are unrelated to the causal explanation of interest \citep{dlp-phd, lake2016building, marcus2018, welling2019}.
Following this line, we leverage tools from causation to develop the mathematics of spurious and invariant correlations, in order to alleviate the excessive reliance of machine learning systems on data biases, allowing them to generalize to new test distributions.

As a thought experiment, consider the problem of classifying images of cows and camels \citep{beery2018recognition}.
To address this task, we label images of both types of animals.
Due to a selection bias, most pictures of cows are taken in green pastures, while most pictures of camels happen to be in deserts.
After training a convolutional neural network on this dataset, we observe that the model fails to classify easy examples of images of cows when they are taken on sandy beaches.
Bewildered, we later realize that our neural network successfully minimized its training error using a simple cheat: classify green landscapes as cows, and beige landscapes as camels.

To solve the problem described above, we need to identify which properties of the training data describe spurious correlations (landscapes and contexts), and which properties represent the phenomenon of interest (animal shapes).
Intuitively, a correlation is spurious when we do not expect it to hold in the future in the same manner as it held in the past.
In other words, spurious correlations do not appear to be stable properties~\citep{woodward2005making}.
Unfortunately, most datasets are not provided in a form amenable to discover stable properties.
Because most machine learning algorithms depend on the assumption that training and testing data are sampled independently from the same distribution~\citep{vapnik-1998}, it is common practice to shuffle at random the training and testing examples.
For instance, whereas the original NIST handwritten data was collected from different writers under different conditions~\citep{grother-1995}, the popular MNIST training and testing sets~\citep{bottou-cortes-94} were carefully shuffled to represent similar mixes of writers.
Shuffling brings the training and testing distributions closer together, but discards what information is stable across writers. 
However, shuffling the data is something that we do, not something that Nature does for us.
When shuffling, we destroy information about how the data distribution changes when one varies the data sources or collection specifics.
Yet, this information is precisely what tells us whether a property of the data is spurious or stable.

Here we take a step back, and assume that the training data is collected into distinct, separate environments.
These could represent different measuring circumstances, locations, times, experimental conditions, external interventions, contexts, and so forth.
Then, we promote learning correlations that are stable across training environments, as these should (under conditions that we will study) also hold in novel testing environments.

Returning to our motivational example, we would like to label pictures of cows and camels under different environments. 
For instance, the pictures of cows taken in the first environment may be located in green pastures 80\pc{} of the time.
In the second environment, this proportion could be slightly different, say 90\pc{} of the time (since pictures were taken in a different country).
These two datasets reveal that ``cow'' and ``green background'' are linked by a strong, but varying (spurious) correlation, which should be discarded in order to generalize to new environments.
Learning machines which pool the data from the two environments together may still rely on the background bias when addressing the prediction task.
But, we believe that all cows exhibit features that allow us to recognize them as so, regardless of their context.

This suggests that invariant descriptions of objects relate to the causal explanation of the object itself (``\emph{Why is it a cow?}'') \citep{lopez2017discovering}.
As shown by \cite{peters2015causal, heinze2018invariant}, there exists an intimate link between invariance and causation useful for generalization.
However, \cite{peters2015causal} assumes a meaningful causal graph relating the observed variables, an awkward assumption when dealing with perceptual inputs such as pixels.
Furthermore, \cite{peters2015causal} only applies to linear models, and scales exponentially with respect to the number of variables in the learning problem.
As such, the seamless integration of causation tools \cite{peters2017elements} into machine learning pipelines remains cumbersome, disallowing what we believe to be a powerful synergy.
Here, we work to address these concerns. 

\paragraph{Contributions} We propose Invariant Risk Minimization (IRM), a novel learning paradigm that estimates nonlinear, invariant, causal predictors from multiple training environments, to enable out-of-distribution (OOD) generalization.
To this end, we first analyze in Section~\ref{sec:robustness} how different learning techniques fail to generalize OOD.
From this analysis, we derive our IRM principle in Section~\ref{sec:invariance}:
\begin{center}
    \emph{To learn invariances across environments, find a data
    representation such that the optimal classifier on top of that representation matches for all environments.}
\end{center}
Section~\ref{sec:causation} examines the fundamental links between causation, invariance, and OOD generalization. 
Section~\ref{sec:experiments} contains basic numerical simulations to validate our claims empirically. 
Section~\ref{sec:outlook} concludes with a Socratic dialogue discussing directions for future research.

\section{The many faces of generalization}
\label{sec:robustness}

Following~\citep{peters2015causal}, we consider datasets $D_e := \{(x^e_i, y^e_i)\}_{i=1}^{n_e}$ collected under multiple training environments $e \in \Etrain$.
These environments describe the same pair of random variables measured under different conditions.
The dataset $D_e$, from environment $e$, contains examples identically and independently distributed according to some probability distribution $P(X^e, Y^e)$.\footnote{We omit the superscripts ``$^e$'' when referring to a random variable regardless of the environment.}
Then, our goal is to use these multiple datasets to learn a predictor $Y \approx f(X)$, which performs well across a large set of unseen but related environments $\Eall \supset \Etrain$.
Namely, we wish to minimize
$$R^{\text{OOD}}(f) = \max_{e \in \Eall} R^e(f)$$
where $R^e(f) := \EE_{X^e, Y^e}[\ell(f(X^e), Y^e)]$ is the risk under environment $e$.
Here, the set of all environments $\Eall$ contains all possible experimental conditions concerning our system of variables, both observable and hypothetical.
This is in the spirit of modal realism and possible worlds \citep{lewis2013counterfactuals}, where we could consider, for instance, environments where we switch off the Sun.
An example clarifies our intentions.
\begin{example}\label{ex:example}
Consider the structural equation model \citep{Wright1921CorrelationAndCausation}:
\begin{align*}
    X_1 &\leftarrow \mathrm{Gaussian}(0, \sigma^2), \nonumber \\
    Y &\leftarrow X_1 + \mathrm{Gaussian}(0, \sigma^2), \\
    X_2 &\leftarrow Y + \mathrm{Gaussian}(0, 1).\nonumber
\end{align*}
\end{example}
As we formalize in Section~\ref{sec:causation}, the set of all environments $\Eall$ contains all modifications of the structural equations for $X_1$ and $X_2$, and those varying the noise of $Y$ within a finite range $[0, \sigma^2_\text{MAX}]$.
For instance, $e \in \Eall$ may replace the equation of $X_2$ by $X^e_2 \leftarrow 10^6$, or vary $\sigma^2$ within this finite range .
To ease exposition consider:
\begin{equation*}
    \Etrain = \{ \text{replace } \sigma^2 \text{ by } 10, \text{ replace } \sigma^2 \text{ by } 20 \}.
\end{equation*}
Then, to predict $Y$ from $(X_1, X_2)$ using a least-squares predictor $\hat{Y}^e = X_1^e\hat{\alpha}_1  +  X_2^e\hat{\alpha}_2$ for environment $e$, we can:
\begin{itemize}
    \item regress from $X_1^e$, to obtain $\hat{\alpha}_1 = 1$ and $\hat{\alpha}_2 = 0$,
    \item regress from $X_2^e$, to obtain $\hat{\alpha}_1 = 0$ and $\hat{\alpha}_2 = \sigma(e)^2 / (\sigma(e)^2 + \frac{1}{2})$,
    \item regress from $(X_1^e, X_2^e)$, to obtain $\hat{\alpha}_1 = 1 / (\sigma(e)^2 + 1)$ and $\hat{\alpha}_2 = \sigma(e)^2 / (\sigma(e)^2 + 1)$.
\end{itemize}
The regression using $X_1$ is our first example of an invariant correlation: this is the only regression whose coefficients do not depend on the environment $e$.
Conversely, the second and third regressions exhibit coefficients that vary from environment to environment.
These varying (spurious) correlations would not generalize well to novel test environments.
Also, not all invariances are interesting:
the regression from the empty set of features into $Y$ is invariant, but of weak predictive power.

The invariant rule $\hat{Y} = 1 \cdot X_1 + 0 \cdot X_2$ is the only predictor with finite $R^\text{OOD}$ across $\Eall$ (to see this, let $X_2 \to \infty$).
Furthermore, this predictor is the causal explanation about how the target variable takes values across environments.
In other words, it provides the correct description about how the target variable reacts in response to interventions on each of the inputs.
This is compelling, as invariance is a statistically testable quantity that we can measure to discover causation.
We elaborate on the relationship between invariance and causation in Section~\ref{sec:causation}.
But first, how can we learn the invariant, causal regression?
Let us review four techniques commonly discussed in prior work, as well as their limitations.

First, we could merge the data from all the training environments and learn a predictor that minimizes the training error across the pooled data, using all features.
This is the ubiquitous Empirical Risk Minimization (ERM) principle \citep{vapnik1992}.
In this example, ERM would grant a large positive coefficient to $X_2$ if the pooled training environments lead to large $\sigma^2(e)$ (as in our example), departing from invariance.

Second, we could minimize $R^\text{rob}(f) = \max_{e \in \Etrain} R^e(f) - r_e$, a robust learning objective where the constants $r_e$ serve as environment baselines \citep{robustsl, ben2009robust, duchi2016statistics, sinha2017certifying}.
Setting these baselines to zero leads to minimizing the maximum error across environments.
Selecting these baselines adequately prevents noisy environments from dominating optimization.
For example, \citet{meinshausen-buhlman-2015} selects $r_e = \mathbb{V}[Y^e]$ to maximize the minimal explained variance across environments.
While promising, robust learning turns out to be equivalent to minimizing a weighted average of environment training errors: 
\begin{proposition}
    \label{prop:robust}
    Given KKT differentiability and qualification conditions, $\exists \lambda_e \geq 0$ such that the minimizer of $R^\mathrm{rob}$ is a first-order stationary point of $\sum_{e \in \Etrain} \lambda_e R^e(f)$.
\end{proposition}
This proposition shows that robust learning and ERM (a special case of robust learning with $\lambda_e = \frac{1}{|\Etrain|}$) would never discover the desired invariance, obtaining infinite $R^{\text{OOD}}$.
This is because minimizing the risk of any mixture of environments associated to large $\sigma^2(e)$ yields a predictor with a large weight on $X_2$.
Unfortunately, this correlation will vanish for testing environments associated to small $\sigma^2(e)$.

Third, we could adopt a domain adaptation strategy, and estimate a data representation $\Phi(X_1, X_2)$ that follows the same distribution for all environments \citep{ganin2016domain, louppe2017learning}.
This would fail to find the true invariance in Example~\ref{ex:example}, since the distribution of the true causal feature $X_1$ (and the one of the target $Y$) can change across environments.
This illustrates why techniques matching feature distributions sometimes attempt to enforce the wrong type of invariance, as discussed in Appendix~\ref{app:ada}.

Fourth, we could follow invariant causal prediction techniques \citep{peters2015causal}.
These search for the subset of variables that, when used to estimate individual regressions for each environment, produce regression residuals with equal distribution across all environments.
Matching residual distributions is unsuited for our example, since the noise variance in $Y$ may change across environments. 

In sum, finding invariant predictors even on simple problems such as Example~\ref{ex:example} is surprisingly difficult.
To address this issue, we propose Invariant Risk Minimization (IRM), a learning paradigm to extract nonlinear invariant predictors across multiple environments, enabling OOD generalization.

\section{Algorithms for invariant risk minimization}
\label{sec:invariance}

In statistical parlance, our goal is to learn correlations invariant across training environments.
For prediction problems, this means finding a data representation such that the optimal classifier,\footnote{We will also use the term ``classifier'' to denote the last layer $w$ for regression problems.} on top of that data representation, is the same for all environments.
More formally:

\begin{definition}
    We say that a data representation $\Phi: \mathcal{X} \to \mathcal{H}$
    elicits an invariant predictor $w \circ \Phi$ 
    across environments $\mathcal{E}$ if there is a classifier 
    $w:\mathcal{H}\to\mathcal{Y}$ simultaneously
    optimal for all environments, that is,
    $w \in \argmin_{\bar w : \mathcal{H} \to \mathcal{Y}} R^e(\bar w \circ \Phi)$
    for all $e \in \mathcal{E}$.
    \label{def:invariant}
\end{definition}

Why is Definition~\ref{def:invariant} equivalent to learning features whose correlations with the target variable are stable?
For loss functions such as the mean squared error and the cross-entropy, optimal classifiers can be written as conditional expectations. 
In these cases, a data representation function~$\Phi$ elicits an invariant predictor across environments $\mathcal{E}$ if and only if for all $h$ in the intersection of the supports of $\Phi(X^e)$ we have
$\EE[Y^e|\Phi(X^e) = h] = \EE[Y^{e'}|\Phi(X^{e'}) = h]$, for all $e, e' \in \mathcal{E}$.

We believe that this concept of invariance clarifies common induction methods in science.
Indeed, some scientific discoveries can be traced to the realization that distinct but potentially related phenomena, once described with the correct variables, appear to obey the same exact physical laws.
The precise conservation of these laws suggests that they remain valid on a far broader range of conditions.
If both Newton's apple and the planets obey the same equations, chances are that gravitation is a thing.

To discover these invariances from empirical data, we introduce Invariant Risk Minimization (IRM), a learning paradigm to estimate data representations eliciting invariant predictors $w \circ \Phi$ across multiple environments.
To this end, recall that we have two goals in mind for the data representation $\Phi$: we want it to be useful to predict well, and elicit an invariant predictor across $\Etr$.
Mathematically, we phrase these goals as the constrained optimization problem:
\begin{equation} \label{eq:irmconst}
\begin{aligned}
    &\min_{\substack{\Phi : \mathcal{X} \to \mathcal{H}\\ w : \mathcal{H} \to \mathcal{Y}}} & & \sum_{e \in \Etrain} R^e(w \circ \Phi) \\
    &\text{subject to} & & w \in \argmin_{\bar w: \mathcal{H} \rightarrow \mathcal{Y}} R^e(\bar w \circ \Phi), \text{ for all $e \in \Etrain$}.
\end{aligned}
\tag{IRM}
\end{equation}
This is a challenging, bi-leveled optimization problem, since each constraint calls an inner optimization routine. 
So, we instantiate \eqref{eq:irmconst} into the practical version: 
\begin{mdframed}[roundcorner=5pt, backgroundcolor=yellow!8]
  \begin{equation} \label{eq:irm1}
  \min_{\Phi: \mathcal{X} \rightarrow \mathcal{Y}} \sum_{e \in \Etrain} R^e(\Phi) + \lambda \cdot \| \nabla_{w|{w = 1.0}}\, R^e(w \cdot \Phi) \|^2,
  \tag{IRMv1}
\end{equation}
\end{mdframed}
where $\Phi$ becomes the entire invariant predictor, $w = 1.0$ is a scalar and fixed ``dummy'' classifier, the gradient norm penalty is used to measure the optimality of the dummy classifier at each environment $e$, and $\lambda \in [0, \infty)$ is a regularizer balancing between predictive power (an ERM term), and the invariance of the predictor $1 \cdot \Phi(x)$.

\subsection{From \eqref{eq:irmconst} to \eqref{eq:irm1}}

This section is a voyage circumventing the subtle optimization issues lurking behind the idealistic objective \eqref{eq:irmconst}, to arrive to the efficient proposal \eqref{eq:irm1}.

\subsubsection{Phrasing the constraints as a penalty}
We translate the hard constraints in~\eqref{eq:irmconst} into the penalized loss 
\begin{equation} \label{eq:irm_a}
  L_{\text{IRM}}(\Phi, w) = \sum_{e \in \Etrain} R^e(w \circ \Phi) + \lambda \cdot \mathbb{D}(w, \Phi, e)
\end{equation}
where
%\begin{equation}
%  w^e_\Phi \in \argmin_{\bar w: \mathcal{H} \rightarrow \mathcal{Y}} R^e(\bar w \circ \Phi)
%  \label{eq:argmin}
%\end{equation}
$\Phi : \mathcal{X} \to \mathcal{H}$, the function $\mathbb{D}(w, \Phi, e)$ measures how close $w$ is to minimizing $R^e(w \circ \Phi)$, and $\lambda \in [0, \infty)$ is a hyper-parameter balancing predictive power and invariance.
In practice, we would like $\mathbb{D}(w, \Phi, e)$ to be differentiable with respect to $\Phi$ and $w$.
Next, we consider linear classifiers $w$ to propose one alternative.

\subsubsection{Choosing a penalty $\mathbb{D}$ for linear classifiers $w$}
Consider learning an invariant predictor $w \circ \Phi$, where $w$ is a linear-least squares regression, and $\Phi$ is a nonlinear data representation. 
In the sequel, all vectors $v \in \RR^d$ are by default in column form, and we denote by $v^\top \in \RR^{1 \times d}$ the row form.
By the normal equations, and given a fixed data representation $\Phi$, we can write $w^e_\Phi \in \argmin_{\bar w} R^e(\bar w \circ \Phi)$ as: 
\begin{equation}
    w^e_{\Phi} =
    \EE_{X^e}\left[\Phi(X^e) \Phi(X^e)^\top \right]^{-1}
    \EE_{X^e, Y^e}\left[\Phi(X^e) Y^e\right],
    \label{eq:lse}
\end{equation}
where we assumed invertibility.
This analytic expression would suggest a simple discrepancy between two linear least-squares classifiers:
\begin{equation}
    \mathbb{D}_{\text{dist}}(w, \Phi, e) = \| w - w^{e}_\Phi\|^2.
    \label{eq:penalty_dist}
\end{equation}

\begin{figure}[t!]
  \centering \includegraphics[width=\linewidth]{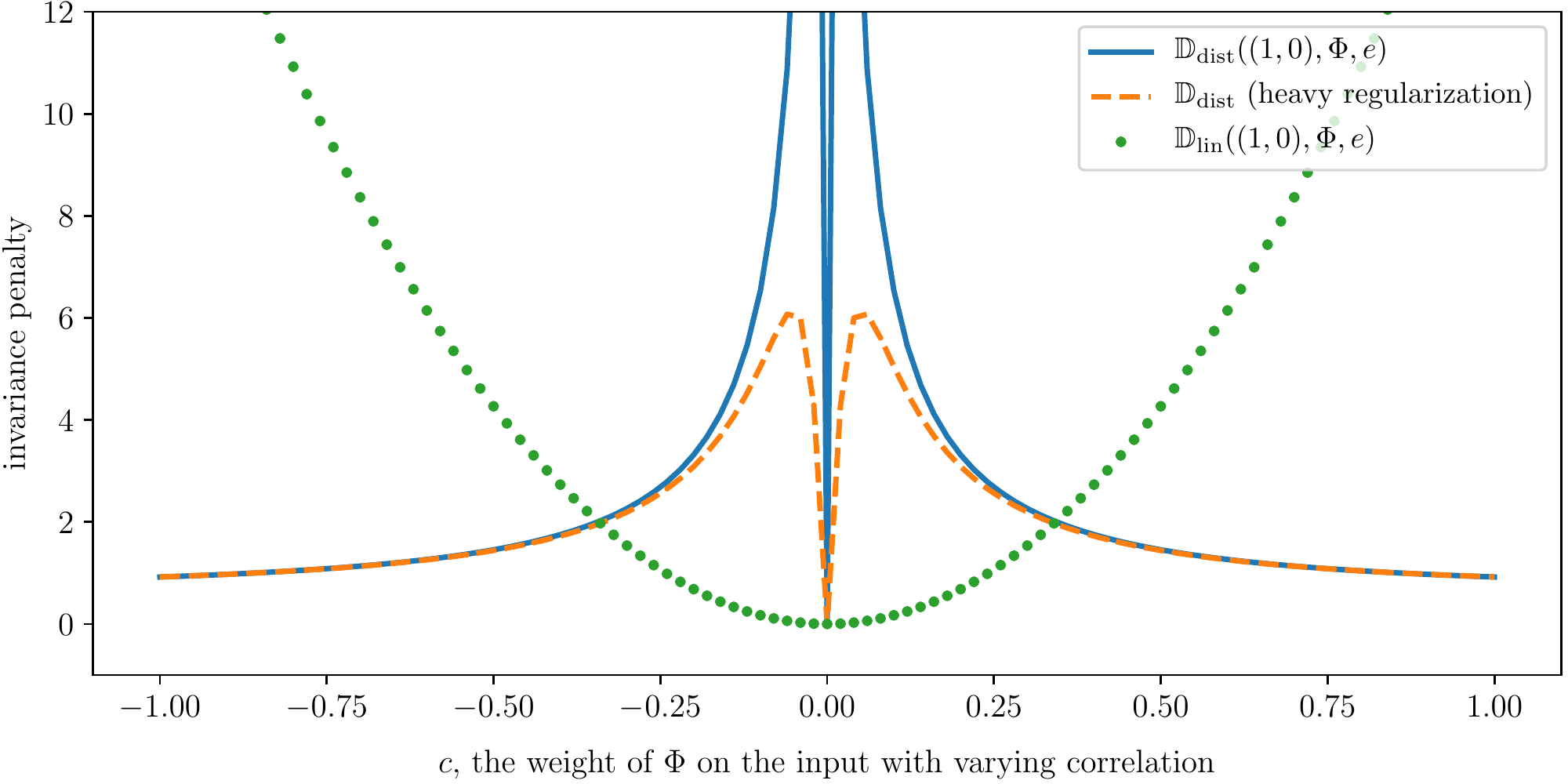} \caption{Different
  measures of invariance lead to different optimization landscapes in
  our Example~\ref{ex:example}.  The na\"ive approach of measuring
  the distance between optimal classifiers $\mathbb{D}_\mathrm{dist}$
  leads to a discontinuous penalty (solid blue unregularized, dashed
  orange regularized).  In contrast, the penalty
  $\mathbb{D}_{\mathrm{lin}}$
  does not exhibit these problems.} \label{fig:penalties}
\end{figure}

Figure~\ref{fig:penalties} uses Example~\ref{ex:example} to show why $\mathbb{D}_{\text{dist}}$ is a poor discrepancy.
The blue curve shows \eqref{eq:penalty_dist} as we vary the coefficient $c$ for a linear data representation $\Phi(x) = x \cdot \mathrm{Diag}([1,c])$, and $w = (1, 0)$.
The coefficient $c$ controls how much the representation depends on the variable $X_2$, responsible for the spurious correlations in Example~\ref{ex:example}. 
We observe that~\eqref{eq:penalty_dist} is discontinuous at $c=0$, the value eliciting the invariant predictor.
This happens because when $c$ approaches zero without being
exactly zero, the least-squares rule \eqref{eq:lse} compensates this change by creating vectors $w_\Phi^e$ whose second coefficient grows to infinity.
This causes a second problem, the penalty approaching zero as~$\|c\| \to \infty$.
The orange curve shows that adding severe regularization to the least-squares regression does not fix these numerical problems.

To circumvent these issues, we can undo the matrix inversion in \eqref{eq:lse} to construct:
\begin{equation}
    \mathbb{D}_{\text{lin}}(w, \Phi, e) =
     \left\| \EE_{X^e}\left[\Phi(X^e) \Phi(X^e)^\top \right]  w - \EE_{X^e, Y^e}\left[\Phi(X^e) Y^e\right]  \right\|^2,
    \label{eq:penalty_ones}
\end{equation}
which measures how much does the classifier $w$ violate the normal equations.
The green curve in Figure~\ref{fig:penalties} shows $\mathbb{D}_\mathrm{\text{lin}}$ as we vary $c$, when setting $w = (1, 0)$.
The penalty $\mathbb{D}_{\text{lin}}$ is smooth (it is a polynomial on both $\Phi$ and $w$), and achieves an easy-to-reach minimum at $c=0$ ---the data representation eliciting the invariant predictor.
Furthermore, $\mathbb{D}_{\text{lin}}(w, \Phi, e) = 0$ if and only if $w \in \argmin_{\bar w} R^e(\bar w \circ \Phi)$.
As a word of caution, we note that the penalty $\mathbb{D}_{\text{lin}}$ is non-convex for general $\Phi$.

\subsubsection{Fixing the linear classifier $w$}

Even when minimizing \eqref{eq:irm_a} over $(\Phi, w)$ using $\mathbb{D}_{\text{lin}}$, we encounter one issue.
When considering a pair $(\gamma \Phi, \frac{1}{\gamma} w)$, it is possible to let $\mathbb{D}_{\text{lin}}$ tend to zero without impacting the ERM term, by letting $\gamma$ tend to zero. 
This problem arises because \eqref{eq:irm_a} is severely over-parametrized.
In particular, for any invertible mapping $\Psi$, we can re-write our invariant predictor as
\begin{equation*}
  w \circ \Phi = \underbrace{\left(w \circ \Psi^{-1}\right)}_{\tilde{w}} \circ \underbrace{\left(\Psi \circ \Phi\right)}_{\tilde{\Phi}}.
\end{equation*}
This means that we can re-parametrize our invariant predictor as to give $w$ any non-zero value $\tilde{w}$ of our choosing.
Thus, we may restrict our search to the data representations for which all the environment optimal classifiers are equal to the same fixed vector $\tilde{w}$.
In words, we are relaxing our recipe for invariance into \emph{finding a data representation such that the optimal classifier, on top of that data representation, is ``$\tilde{w}$'' for all environments}.
This turns~\eqref{eq:irm_a} into a relaxed version of IRM, where optimization only happens over $\Phi$:
\begin{equation} \label{eq:irm_b}
    L_{\mathrm{IRM}, w=\tilde w}(\Phi) = \sum_{e \in \Etrain} R^e(\tilde{w} \circ \Phi) + \lambda \cdot \mathbb{D}_\text{lin}(\tilde{w}, \Phi, e).
\end{equation}
As $\lambda \to \infty$, solutions $(\Phi^*_\lambda, \tilde w)$ of \eqref{eq:irm_b} tend to solutions $(\Phi^*, \tilde w)$ of \eqref{eq:irmconst} for linear $\tilde{w}$.

\subsubsection{Scalar fixed classifiers $\tilde{w}$ are sufficient to monitor invariance}

Perhaps surprisingly, the previous section suggests that $\tilde{w} =
(1, 0, \ldots, 0)$ would be a valid choice for our fixed classifier.
In this case, only the first component of the data representation
would matter!  We illustrate this apparent paradox by providing a
complete characterization for the case of \emph{linear} invariant
predictors. In the following theorem, matrix $\Phi\in\RR^{p\times d}$
parametrizes the data representation function, vector $w\in\RR^p$ the
simultaneously optimal classifier, and $v=\Phi^\top w$ the
predictor $w\circ\Phi$.

\begin{theorem}
  \label{thm:solutions} For all $e\in\mathcal{E}$, let
  ${R^e:\R^d\to\R}$ be convex differentiable cost functions.
  A vector $v\in\RR^d$ can be written $v=\Phi^\top w$,
  where $\Phi\in\R^{p \times d}$, and where $w\in\R^{p}$ simultaneously
  minimize $R^e(w\circ\Phi)$ for all $e\in\mathcal{E}$, if and only
  if $v^\top \nabla {R^e(v)}=0$ for all $e\in\mathcal{E}$.
  Furthermore, the matrices $\Phi$ for which such a decomposition exists
  are the matrices whose nullspace $\mathrm{Ker}(\Phi)$ is
  orthogonal to $v$ and contains all the~$\nabla R^e(v)$.
\end{theorem}

So, any linear invariant predictor can be decomposed as linear data representations of different ranks. 
In particular, we can restrict our search to matrices $\Phi \in \R^{1\times d}$ and let $\tilde{w}\in\RR^1$
be the fixed scalar $1.0$.
This translates~\eqref{eq:irm_b} into:
\begin{equation} \label{eq:irm_c}
    L_{\text{IRM}, w=1.0}(\Phi^\top) = \sum_{e \in \Etrain} R^e(\Phi^\top) + \lambda \cdot \mathbb{D}_{\text{lin}}(1.0, \Phi^\top, e).
\end{equation}
Section~\ref{sec:causation} shows that the existence of decompositions with high-rank data representation matrices $\Phi^\top$ are key to out-of-distribution generalization, regardless of whether we restrict IRM to search for rank-1 $\Phi^\top$.

\begin{figure}
  \centering \includegraphics[]{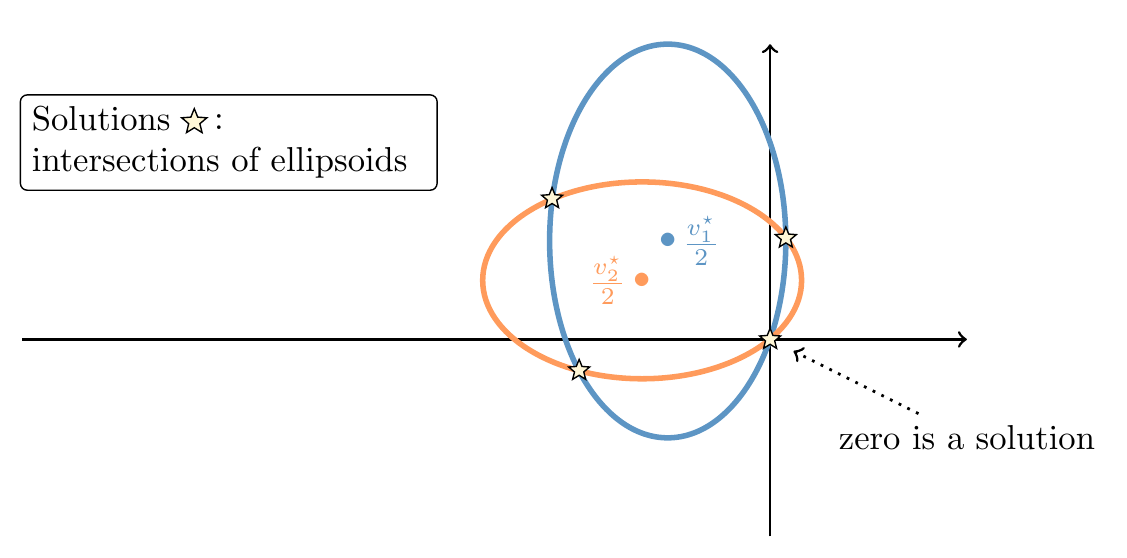}
  \caption{\label{fig:ellipsoids} The solutions of the invariant
    linear predictors $v = \Phi^\top w$ coincide with the intersection of
    the ellipsoids representing the orthogonality condition
    $v^\top\nabla{R^e(v)}=0$.}
\end{figure}

Geometrically, each orthogonality condition $v^\top \nabla R^e(v) = 0$ in Theorem~\ref{thm:solutions} defines a~$(d{-}1)$-dimensional manifold in~$\R^d$.
Their intersection is itself a manifold of
dimension greater than $d{-}m$, where $m$ is the number of
environments.
When using the squared loss, each condition is a
quadratic equation whose solutions form an ellipsoid in
$\R^d$. Figure~\ref{fig:ellipsoids} shows how their intersection is
composed of multiple connected components, one of which contains the
trivial solution $v=0$. This shows that~\eqref{eq:irm_c} remains
nonconvex, and therefore sensitive to initialization.

\subsubsection{Extending to general losses and multivariate outputs}
Continuing from~\eqref{eq:irm_c}, we obtain our final algorithm~\eqref{eq:irm1} by realizing that the invariance penalty~\eqref{eq:penalty_ones}, introduced for the least-squares case, can be written as a general function of the risk, namely $\mathbb{D}(1.0, \Phi, e) = \| \nabla_{w|w=1.0} R^e(w \cdot \Phi) \|^2$, where $\Phi$ is again a possibly nonlinear data representation.
This expression measures the optimality of the fixed scalar classifier $w = 1.0$ for any convex loss, such as the cross-entropy.
If the target space $\mathcal{Y}$ returned by $\Phi$ has multiple outputs, we multiply all of them by the fixed scalar classifier $w = 1.0$.

\subsection{Implementation details}
\label{sub:implementation_details}

When estimating the objective \eqref{eq:irm1} using mini-batches for stochastic gradient descent,
one can obtain an unbiased estimate of the squared gradient norm as
\begin{equation*}
    \sum_{k=1}^b \left[ \nabla_{w|w=1.0} \ell(w \cdot \Phi(X^{e, i}_k), Y^{e, i}_k) \cdot \nabla_{w|w=1.0} \ell(w \cdot \Phi(X^{e, j}_k), Y^{e,j}_k) \right],
\end{equation*}
where $(X^{e, i}, Y^{e, i})$ and $(X^{e, j}, Y^{e, j})$ are two random mini-batches of size $b$ from environment $e$, and $\ell$ is a loss function.
We offer a PyTorch example in Appendix~\ref{sec:code}.

\subsection{About nonlinear invariances $w$}
\label{sub:nonlinw}

How restrictive is it to assume that the invariant optimal classifier $w$ is linear?
One may argue that given a sufficiently flexible data representation $\Phi$, it is possible to write any invariant predictor as $1.0 \cdot \Phi$.
However, enforcing a linear invariance may grant non-invariant predictors a penalty $\mathbb{D}_\text{lin}$ equal to zero.
For instance, the null data representation $\Phi_0(X^e) = 0$ admits any $w$ as optimal amongst all the \emph{linear} classifiers for all environments.
But, the elicited predictor $w \circ \Phi_0$ is not invariant in cases where $\EE[Y^e] \neq 0$.
Such null predictor would be discarded by the ERM term in the IRM objective.
In general, minimizing the ERM term $R^e(\tilde w \circ \Phi)$ will drive $\Phi$ so that $\tilde w$ is optimal amongst all predictors, even if $\tilde w$ is linear.

We leave for future work several questions related to this issue.
Are there non-invariant predictors that would not be discarded by either the ERM or the invariance term in IRM?
What are the benefits of enforcing non-linear invariances $w$ belonging to larger hypothesis classes $\mathcal{W}$? 
How can we construct invariance penalties $\mathbb{D}$ for non-linear invariances?

\section{Invariance, causality and generalization}
\label{sec:causation}

The newly introduced IRM principle promotes low error and invariance across training environments $\Etr$.
When do these conditions imply invariance across all environments $\Eall$?
More importantly, when do these conditions lead to low error across $\Eall$, and consequently out-of-distribution generalization?
And at a more fundamental level, how does statistical invariance and out-of-distribution generalization relate to concepts from the theory of causation? 

So far, we have omitted how different environments should relate to enable out-of-distribution generalization.
The answer to this question is rooted in the theory of causation.
We begin by assuming that the data from all the environments share the same underlying Structural Equation Model, or SEM \citep{Wright1921CorrelationAndCausation, pearl2009causation}:

\begin{definition}
    A \emph{Structural Equation Model (SEM)} $\matC := (\matS, N)$ governing the random vector $X = (X_1, \ldots, X_d)$ is a set of \emph{structural equations}:
    \begin{equation*}
        \matS_i : X_i \leftarrow f_i(\PA(X_i), N_i),
    \end{equation*}
    where $\PA(X_i) \subseteq \{ X_1, \ldots, X_d \} \setminus \{ X_i \}$ are called the parents of $X_i$, and the $N_i$ are independent noise random variables.
    We say that ``$X_i$ causes $X_j$'' if $X_i \in \PA(X_j)$.
    We call \emph{causal graph of $X$} to the graph obtained by drawing i) one node for each $X_i$, and ii) one edge from $X_i$ to $X_j$ if $X_i \in \PA(X_j)$.
    We assume acyclic causal graphs.
\end{definition}

By running the structural equations of a SEM $\matC$ according to the topological ordering of its causal graph, we can draw samples from the \emph{observational distribution $P(X)$}.
In addition, we can manipulate (intervene) an unique SEM in different ways, indexed by $e$, to obtain different but related SEMs $\mathcal{C}^e$. 

\begin{definition} Consider a SEM $\matC = (\matS, N)$. An \emph{intervention $e$ on $\matC$} consists of replacing one or several of its structural equations to obtain an intervened SEM $\matC^e = (\matS^e, N^e)$, with structural equations:
\begin{equation*}
    S^e_i : X^e_i \leftarrow f^e_i(\PA^e(X_i^e), N^e_i),
\end{equation*}
The variable $X^e$ is \emph{intervened} if $S_i \neq S^e_i$ or $N_i \neq N^e_i$.
\end{definition}

Similarly, by running the structural equations of the intervened SEM $\matC^e$, we can draw samples from the \emph{interventional distribution $P(X^e)$}.
For instance, we may consider Example~\ref{ex:example} and intervene on $X_2$, by holding it constant to zero, thus replacing the structural equation of $X_2$ by $X^e_2 \leftarrow 0$.
Admitting a slight abuse of notation, each intervention $e$ generates a new environment $e$ with interventional distribution $P(X^e, Y^e)$.
\emph{Valid interventions} $e$, those that do not destroy too much information about the target variable $Y$, form the set of all environments $\Eall$. 

Prior work \citep{peters2015causal} considered valid interventions as those that do not change the structural equation of $Y$, since arbitrary interventions on this equation render prediction impossible.
In this work, we also allow changes in the noise variance of $Y$, since varying noise levels appear in real problems, and these do not affect the optimal prediction rule.
We formalize this as follows.
\begin{definition} Consider a SEM $\matC$ governing the random vector $(X_1, \ldots, X_d, Y)$, and the learning goal of predicting $Y$ from $X$. 
  Then, the \emph{set of all environments $\Eall(\matC)$} indexes all the interventional distributions $P(X^e, Y^e)$ obtainable by \emph{valid interventions} $e$.
    An {intervention $e \in \Eall(\matC)$} is valid as long as (i) the causal graph remains acyclic, (ii) $\EE[Y^e | \PA(Y)] = \EE[Y | \PA(Y)]$, and (iii) $\mathbb{V}[Y^e | \PA(Y)]$ remains within a finite range.
\end{definition}

Condition (iii) can be waived if one takes into account environment specific baselines into the definition of $R^{\mathrm{OOD}}$, similar to those appearing in the robust learning objective $R^{\mathrm{rob}}$. We leave additional quantifications of out-of-distribution generalization for future work. 

The previous definitions establish fundamental links between causation and invariance.
Moreover, one can show that a predictor $v: \mathcal{X} \rightarrow \mathcal{Y}$ is invariant across $\Eall(\mathcal{C})$ if and only if it attains optimal $R^{\text{OOD}}$, and if and only if it uses only the direct causal parents of $Y$ to predict, that is, $v(x) = \EE_{N_Y}\left[f_Y({\PA(Y)}, N_Y)\right]$.
The rest of this section follows on these ideas to showcase how invariance across training environments can enable out-of-distribution generalization across all environments.

\subsection{Generalization theory for IRM}

The goal of IRM is to build predictors that generalize out-of-distribution, that is, achieving low error across $\Eall$.
To this end, IRM enforces low error and invariance across $\Etrain$.
The bridge from low error and invariance across $\Etrain$ to low error across $\Eall$ can be traversed in two steps. 

First, one can show that low error across $\Etrain$ and invariance across $\Eall$ leads to low error across $\Eall$.
This is because, once the data representation $\Phi$ eliciting an invariant predictor $w \circ \Phi$ across $\Eall$ is estimated, the generalization error of $w \circ \Phi$ respects standard error bounds. 
Second, we examine the remaining condition towards low error across $\Eall$:
namely, under which conditions does invariance across training environments $\Etrain$ imply invariance across all environments $\Eall$?
%
%We explore this question for \emph{linear} IRM.

For linear IRM, our starting point to answer this question is the theory of Invariant Causal Prediction (ICP) \citep[Theorem 2]{peters2015causal}.
There, the authors prove that ICP recovers the target invariance as long as the data (i) is Gaussian, (ii) satisfies a linear SEM, and (iii) is obtained by certain types of interventions. 
Theorem~\ref{theo:lingen} shows that IRM learns such invariances even when these three assumptions fail to hold.
In particular, we allow for non-Gaussian data, dealing with observations produced as a linear transformation of the variables with stable and spurious correlations, and do not require specific types of interventions or the existence of a causal graph.

The setting of the theorem is as follows.
$Y^e$ has an invariant correlation with an unobserved latent variable $Z_1^e$ by a linear relationship $Y^e = Z_1^e \cdot \gamma + \epsilon^e$, with $\epsilon^e$ independent of $Z_1^e$.
What we observe is $X^e$, which is a scrambled combination of $Z_1^e$ and another variable $Z_2^e$ that can be arbitrarily correlated with $Z_1^e$ and $\epsilon^e$.
Simply regressing using all of $X^e$ will then recklessly exploit $Z_2^e$ (since it gives extra, but spurious, information on $\epsilon^e$ and thus $Y^e$).
A particular instance of this setting is when $Z_1^e$ is the cause of $Y^e$, $Z_2^e$ is an effect, and $X^e$ contains both causes and effects.
To generalize out of distribution the representation has to discard $Z_2^e$ and keep $Z_1^e$.

Before showing Theorem~\ref{theo:lingen}, we need to make our assumptions precise.
To learn useful invariances, one must require some degree of diversity across training environments.
On the one hand, extracting two random subsets of examples from a large dataset does not lead to diverse environments, as both subsets would follow the same distribution.
On the other hand, splitting a large dataset by conditioning on arbitrary variables can generate diverse environments, but may introduce spurious correlations and destroy the invariance of interest \citep[Section 3.3]{peters2015causal}.
Therefore, we will require sets of training environments containing sufficient diversity and satisfying an underlying invariance.
We formalize the diversity requirement as needing envirnments to lie in \emph{linear general position}.

\begin{assu}\label{a:gpos}
    A set of training environments $\Etrain$ lie in \emph{linear general position} of degree $r$ if
    $|\Etrain| > d - r + \frac{d}{r}$ for some $r \in \NN$,
    and for all non-zero $x \in \RR^{d}$:
    $$\dim\left(\mathrm{span}\left(\left\{\EE_{X^e}\left[{X^e} {X^e}^\top\right] x - \EE_{X^e, \epsilon^e}\left[{X^e} \epsilon^e\right]\right\}_{e \in \Etrain} \right)\right) > d - r.$$
\end{assu}
Intuitively, the assumption of linear general position limits the extent to which the training environments are co-linear.
Each new environment laying in linear general position will remove one degree of freedom in the space of invariant solutions.
Fortunately, Theorem~\ref{theo:genp1} shows that the set of cross-products $\EE_{X^e}[{X^e} {X^e}^\top]$ not satisfying a linear general position has measure zero.
Using the assumption of linear general position, we can show that the invariances that IRM learns across training environments transfer to all environments.

In words, the next theorem states the following.
If one finds a representation $\Phi$ of rank $r$ eliciting an invariant predictor $w \circ \Phi$ across $\Etrain$, and $\Etrain$ lie in linear general position of degree $r$, then $w \circ \Phi$ is invariant across $\Eall$.

\begin{theorem} \label{theo:lingen}
    Assume that
    \begin{align*}
        Y^e   &= Z_1^e \cdot \gamma + \epsilon^e, \quad Z^e_1 \perp \epsilon^e, \quad \EE[\epsilon^e] = 0,\\
        X^e   &= S(Z^e_1, Z^e_2).
    \end{align*}
    Here, $\gamma \in \RR^c$, $Z_1^e$ takes values in $\mathbb{R}^{c}$, $Z_2^e$ takes values in $\mathbb{R}^{q}$,
      and $S \in \mathbb{R}^{d \times (c + q)}$.
    Assume that the $Z_1$ component of $S$ is invertible:
      that there exists $\tilde{S} \in \mathbb{R}^{c \times d}$ such that $\tilde{S} \left(S (z_1, z_2)\right) = z_1$, for all $z_1 \in \RR^c, z_2 \in \RR^q$.
    Let $\Phi\in \RR^{d \times d}$ have rank $r > 0$.
    Then, if at least $d - r + \frac{d}{r}$ training environments $\Etrain \subseteq \Eall$ lie in linear general position of degree $r$, we have that
    \begin{equation}
      \Phi\, \EE_{X^e} \left[{X^e} {X^e}^\top \right] \Phi^\top w = \Phi\, \EE_{X^e, Y^e}\left[{X^e} Y^e\right]
        \label{eq:theo_cond}
    \end{equation}
    holds for all $e \in \Etrain$ iff $\Phi$ elicits the invariant predictor $\Phi^\top w$ for all $e \in \Eall$.
\end{theorem}

The assumptions about linearity, centered noise, and independence between the noise $\epsilon^e$ and the causal variables $Z_1$ from Theorem~\ref{theo:lingen} also appear in ICP \citep[Assumption 1]{peters2015causal}, implying the invariance $\EE[Y^e | Z^e_1 = z_1] = z_1 \cdot \gamma$.
As in ICP, we allow correlations between $\epsilon^e$ and the non-causal variables $Z^e_2$, which leads ERM into absorbing spurious correlations (as in our Example~\ref{ex:example}, where $S = I$ and $Z^e_2 = X^e_2$).

In addition, our result contains several novelties. 
First, we do not assume that the data is Gaussian, the existence of a causal graph, or that the training environments arise from specific types of interventions.
Second, the result extends to ``scrambled setups'' where $S \neq I$.
These are situations where the causal relations are not defined on the observable features $X$, but on a latent variable $(Z_1, Z_2)$ that IRM needs to recover and filter. 
Third, we show that representations $\Phi$ with higher rank need fewer training environments to generalize.
This is encouraging, as representations with higher rank destroy less information about the learning problem at hand.

We close this section with two important observations.
First, while robust learning generalizes across interpolations of training environments (recall Proposition~\ref{prop:robust}), learning invariances with IRM buys extrapolation powers.
We can observe this in Example~\ref{ex:example} where, using two training environments, robust learning yields predictors that work well for $\sigma \in [10, 20]$, while IRM yields predictors that work well for all $\sigma$.
Finally, IRM is a differentiable function with respect to the covariances of the training environments.
Therefore, in cases when the data follows an approximately invariant model, IRM should return an approximately invariant solution, being robust to mild model misspecification.
This is in contrast to common causal discovery methods based on thresholding statistical hypothesis tests.

\subsection{On the nonlinear case and the number of environments}

In the same vein as the linear case, we could attempt to provide IRM with guarantees for the nonlinear regime.
Namely, we could assume that each constraint $\| \nabla_{w|w=1.0} R^e(w \cdot \Phi) \| = 0$ removes one degree of freedom from the possible set of solutions $\Phi$.
Then, for a sufficiently large number of diverse training environments, we would elicit the invariant predictor.
Unfortunately, we were unable to phrase such a ``nonlinear general position'' assumption and prove that it holds almost everywhere, as we did in Theorem~\ref{theo:genp1} for the linear case.
We leave this effort for future work.

While general, Theorem~\ref{theo:lingen} is pessimistic, since it requires the number of training environments to scale linearly with the number of parameters in the representation matrix $\Phi$. 
Fortunately, as we will observe in our experiments from Section~\ref{sec:experiments}, it is often the case that two environments are sufficient to recover invariances.
We believe that these are problems where $\EE[Y^e | \Phi(X^e)]$ cannot match for two different environments $e \neq e'$ unless $\Phi$ extracts the causal invariance.
The discussion from Section~\ref{sub:nonlinw} gains relevance here, since enforcing $\mathcal{W}$-invariance for larger families $\mathcal{W}$ should allow discarding more non-invariant predictors with fewer training environments.
All in all, studying what problems allow the discovery of invariances from few environments is a promising line of work towards a learning theory of invariance.

\subsection{Causation as invariance}

We promote invariance as the main feature of causation.
Unsurprisingly, we are not pioneers in doing so.
To predict the outcome of an intervention, we rely on (i) the properties of our intervention and (ii) the properties assumed invariant after the intervention.
Pearl's do-calculus \citet{pearl2009causation} on causal graphs is a framework that tells which conditionals remain invariant after an intervention.
Rubin's ignorability \citet{rubin1974estimating} plays the same role.
What's often described as autonomy of causal mechanisms \citep{haavelmo1944probability, aldrich1989autonomy} is a specification of invariance under intervention.
A large body of philosophical work \citep{skyrms1980causal, redhead1987incompleteness, mitchell2000dimensions, cartwright2003two, woodward2005making, cheng2017causal} studies the close link between invariance and causation.
Some works in machine learning \citep{scholkopf2012causal, ghassami2017learning, heinze2017conditional, kuang2018stable, meinshausen2018causality, rojas2018invariant, magliacane2018domain, bengio2019meta} pursue similar questions.

The invariance view of causation transcends some of the difficulties of working with causal graphs.
For instance, the ideal gas law $PV = nRT$ or Newton's universal gravitation $F = G\frac{m_1 m_2}{r^2}$ are difficult to describe using structural equation models (\emph{What causes what?}), but are prominent examples of laws that are invariant across experimental conditions.
When collecting data about gases or celestial bodies, the universality of these laws will manifest as invariant correlations, which will sponsor valid predictions across environments, as well as the conception of scientific theories.

Another motivation supporting the invariance view of causation are the problems studied in machine learning.
For instance, consider the task of image classification.
Here, the observed variables are hundreds of thousands of correlated pixels.
What is the causal graph governing them?
It is reasonable to assume that causation does not happen between pixels, but between the real-world concepts captured by the camera.
In these cases, invariant correlations in images are a proxy into the causation at play in the real world. 
To find those invariant correlations, we need methods which can disentangle the observed pixels into latent variables closer to the realm of causation, such as IRM.
In rare occasions we are truly interested in the entire causal graph governing all the variables in our learning problem.
Rather, our focus is often on the causal invariances improving generalization across novel distributions of examples. 

\section{Experiments}
\label{sec:experiments}

We perform two experiments to assess the generalization abilities of IRM across multiple environments.
The source-code is available at\\ \url{https://github.com/facebookresearch/InvariantRiskMinimization}.

\subsection{Synthetic data}

As a first experiment, we extend our motivating Example~\ref{ex:example}.
First, we increase the dimensionality of each of the two input features in $X = (X_1, X_2)$ to $10$ dimensions.
Second, as a form of model misspecification, we allow the existence of a $10$-dimensional hidden confounder variable $H$.
Third, in some cases the features $Z$ will not be directly observed, but only a scrambled version $X = S (Z)$.
Figure~\ref{fig:chains_definitions} summarizes the SEM generating the data $(X^e, Y^e)$ for all environments $e$ in these experiments.
More specifically, for environment $e \in \mathbb{R}$, we consider the following variations:
\begin{itemize}
    \item \emph{Scrambled} (S) observations, where $S$ is an orthogonal matrix, or\\
          \emph{unscrambled} (U) observations, where $S = I$.
    \item \emph{Fully-observed} (F) graphs, where $W_{h \to 1} = W_{h \to y} = W_{h \to 2} = 0$, or\\
          \emph{partially-observed} (P) graphs, where $(W_{h \to 1}, W_{h \to y}, W_{h \to 2})$ are Gaussian.
    \item \emph{Homoskedastic} (O) $Y$-noise, where $\sigma_y^2 = e^2$ and $\sigma_2^2 = 1$, or\\
          \emph{heteroskedastic} (E) $Y$-noise, where $\sigma_y^2 = 1$ and $\sigma_2^2 = e^2$.
\end{itemize}

\begin{figure}[ht]
  \centering
  \begin{subfigure}[b]{0.32\linewidth}
    \begin{tikzpicture}[]
        \node[draw=black, thick, circle] (H) at (0, 1.5) {$H^e$};
        \node[draw=black, thick, circle] (X) at (-1.5, 0) {$Z_1^e$};
        \node[draw=black, thick, circle] (Y) at (0, 0) {$Y^e$};
        \node[draw=black, thick, circle] (Z) at (1.5, 0) {$Z_2^e$};
        \draw [->, thick, shorten <= 2pt, shorten >= 2pt] (H) -- (X);
        \draw [->, thick, shorten <= 2pt, shorten >= 2pt] (H) -- (Y);
        \draw [->, thick, shorten <= 2pt, shorten >= 2pt] (H) -- (Z);
        \draw [->, thick, shorten <= 2pt, shorten >= 2pt] (X) -- (Y);
        \draw [->, thick, shorten <= 2pt, shorten >= 2pt] (Y) -- (Z);
    \end{tikzpicture}
  \end{subfigure}
  \begin{subfigure}[b]{0.6\linewidth}
    \begin{align*}
        H^e &\leftarrow \mathcal{N}(0, e^2)\\
        Z_1^e &\leftarrow\mathcal{N}(0, e^2) + W_{h \to 1} H^e \\
        Y^e &\leftarrow Z_1^e \cdot W_{1 \to y}+ \mathcal{N}(0, \sigma^2_y) + W_{h \to y} H^e  \\
        Z_2^e &\leftarrow  W_{y \to 2} Y^e + \mathcal{N}(0, \sigma^2_2) + W_{h \to 2} H^e
    \end{align*}
  \end{subfigure}
    \caption{In our synthetic experiments, the task is to predict $Y^e$ from $X^e = S(Z^e_1, Z^e_2)$. }
%    across multiple environments $e \in \mathbb{R}$.}
\label{fig:chains_definitions}
\end{figure}

These variations lead to eight setups referred to by their initials.
For instance, the setup ``FOS'' considers fully-observed (F), homoskedastic $Y$-noise (O), and scrambled observations (S).
For all variants, $(W_{1 \to y}, W_{y \to 2})$ have Gaussian entries.
Each experiment draws $1000$ samples from the three training environments $\Etrain =  \{0.2, 2, 5\}$.
IRM follows the variant \eqref{eq:irm1}, and uses the environment $e=5$ to cross-validate the invariance regularizer $\lambda$.
We compare to ERM and ICP \citep{peters2015causal}.

\begin{figure}[]
    \centering
    \includegraphics[width=\linewidth]{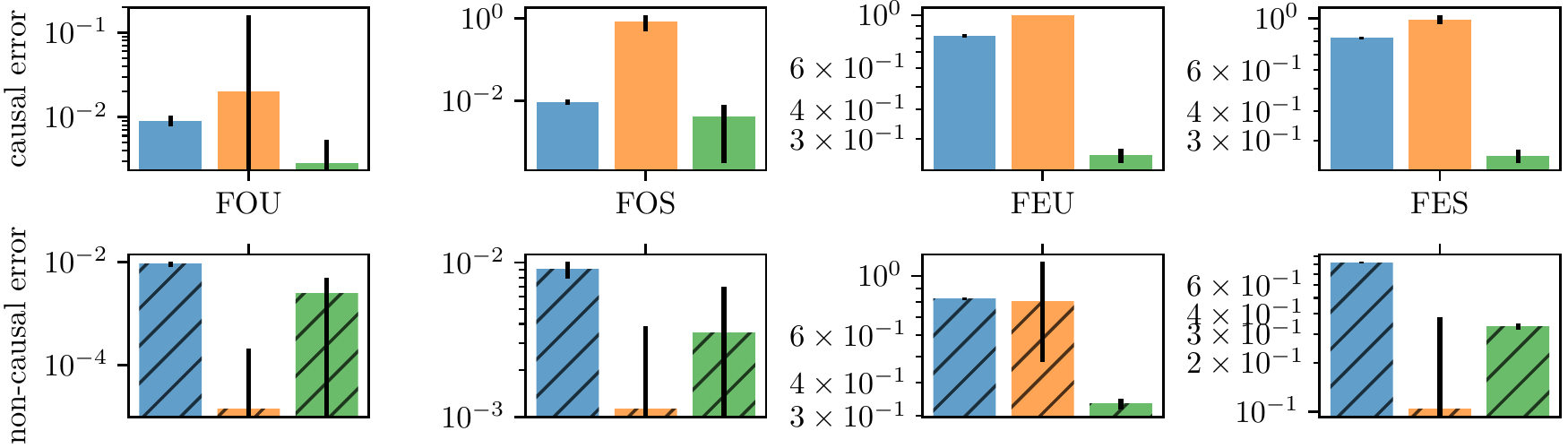}
    \vskip 0.5cm
    \includegraphics[width=\linewidth]{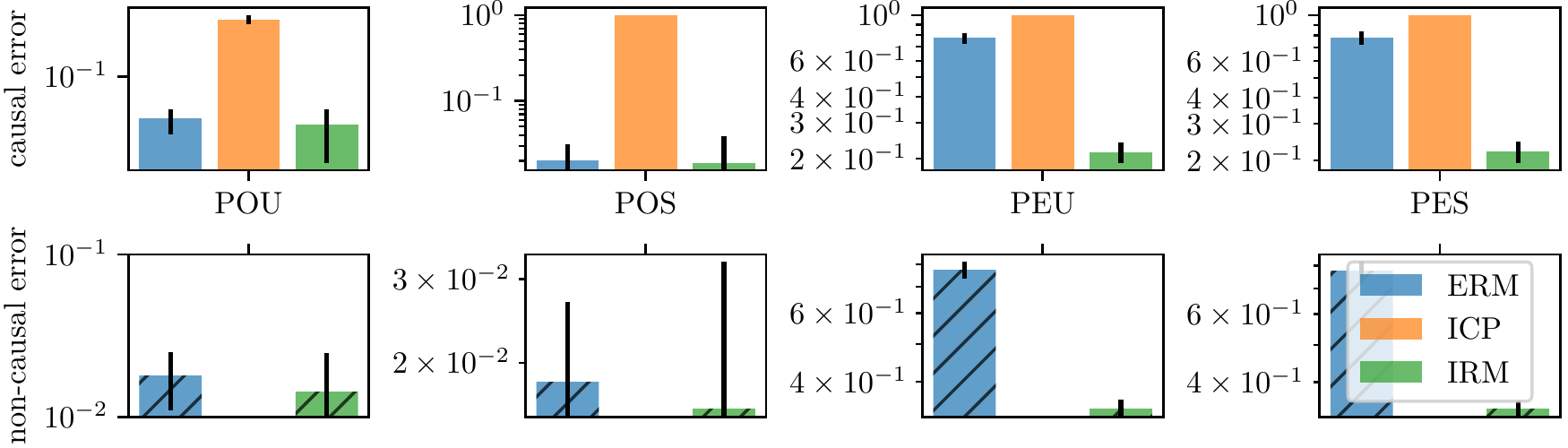}
    \caption{Average errors on causal (plain bars) and non-causal (striped bars) weights for our synthetic experiments. The $y$-axes are in log-scale. See main text for details.}
    \label{fig:chains_results}
\end{figure}

Figure~\ref{fig:chains_results} summarizes the results of our experiments.
We show two metrics for each estimated prediction rule $\hat{Y} = X_1 \cdot \hat{W}_{1 \to y} + X_2 \cdot \hat{W}_{y \to 2}$.
To this end, we consider a de-scrambled version of the estimated coefficients $(\hat{M}_{1 \to y}, \hat{M}_{y \to 2}) = (\hat{W}_{1 \to y}, \hat{W}_{y \to 2})^\top S^\top$.
First, the plain barplots shows the average squared error between $\hat{M}_{1 \to y}$ and $W_{1 \to y}$.
This measures how well does a predictor recover the weights associated to the causal variables.
Second, each striped barplot shows the norm of estimated weights $\hat{M}_{y \to 2}$ associated to the non-causal variable.
We would like this norm to be zero, as the desired invariant causal predictor is $\hat Y^e = (W_{1 \to y}, 0)^\top S^\top(X_1^e, X_2^e)$.
In summary, IRM is able to estimate the most accurate causal and non-causal weights across all experimental conditions.
In most cases, IRM is orders of magnitude more accurate than ERM (our $y$-axes are in log-scale).
IRM also out-performs ICP, the previous state-of-the-art method, by a large margin. 
Our experiments also show the conservative behaviour of ICP (preferring to reject most covariates as direct causes), leading to large errors on causal weights and small errors on non-causal weights.

\subsection{Colored MNIST}
\label{sec:color_mnist}

We validate IRM at learning nonlinear invariant predictors with a synthetic binary classification task derived from MNIST.
The goal is to predict a binary label assigned to each image based on the digit.
Whereas MNIST images are grayscale, we color each image either red or green in a way that correlates strongly (but spuriously) with the class label.
By construction, the label is more strongly correlated with the color than with the digit, so any algorithm purely minimizing training error will tend to exploit the color.
Such algorithms will fail at test time because the direction of the correlation is reversed in the test environment.
By observing that the strength of the correlation between color and label varies between the two training environments, we can hope to eliminate color as a predictive feature, resulting in better generalization.

We define three environments (two training, one test) from MNIST transforming each example as follows:
first, assign a preliminary binary label $\tilde y$ to the image based on the digit: $\tilde y=0$ for digits 0-4 and $\tilde y=1$ for 5-9.
Second, obtain the final label $y$ by flipping $\tilde y$ with probability 0.25.
Third, sample the color id $z$ by flipping $y$ with probability $p^e$, where $p^e$ is 0.2 in the first
environment, 0.1 in the second, and 0.9 in the test one. Finally, color the image red if $z = 1$ or green if $z = 0$.

We train MLPs on the colored MNIST training environments using different objectives and report results in \autoref{tab:colored-mnist-results}.
For each result we report the mean and standard deviation across ten runs.
Training with ERM returns a model with high accuracy in the training environments but below-chance accuracy in the test environment, since the ERM model classifies mainly based on color.
Training with IRM results in a model that performs worse on the training environments, but relies less on the color and hence generalizes better to the test environments.
An oracle that ignores color information by construction outperforms IRM only slightly.

\begin{table}[t!]
\centering
\begin{tabular}{ l r r }
  \toprule
  \textbf{Algorithm} & \textbf{Acc. train envs.} & \textbf{Acc. test env.} \\
  \midrule
  ERM & $87.4 \pm 0.2$ & $17.1 \pm 0.6$ \\
  \textbf{IRM (ours)} & $70.8 \pm 0.9$ & $\mathbf{66.9 \pm 2.5}$ \\
  \midrule
  Random guessing (hypothetical) & 50 & 50 \\
  Optimal invariant model (hypothetical) & 75 & 75 \\
  ERM, grayscale model (oracle) & $73.5 \pm 0.2$ & $73.0 \pm 0.4$ \\
  \bottomrule
\end{tabular}
\caption{Accuracy (\pc{}) of different algorithms on the Colored MNIST synthetic task.
         ERM fails in the test environment because it relies on spurious color correlations to classify digits.
         IRM detects that the color has a spurious correlation with the label and thus uses only the digit to predict,
         obtaining better generalization to the new unseen test environment.}
\label{tab:colored-mnist-results}
\end{table}
\begin{figure}[t!]
  \centering
  \includegraphics[width=\linewidth]{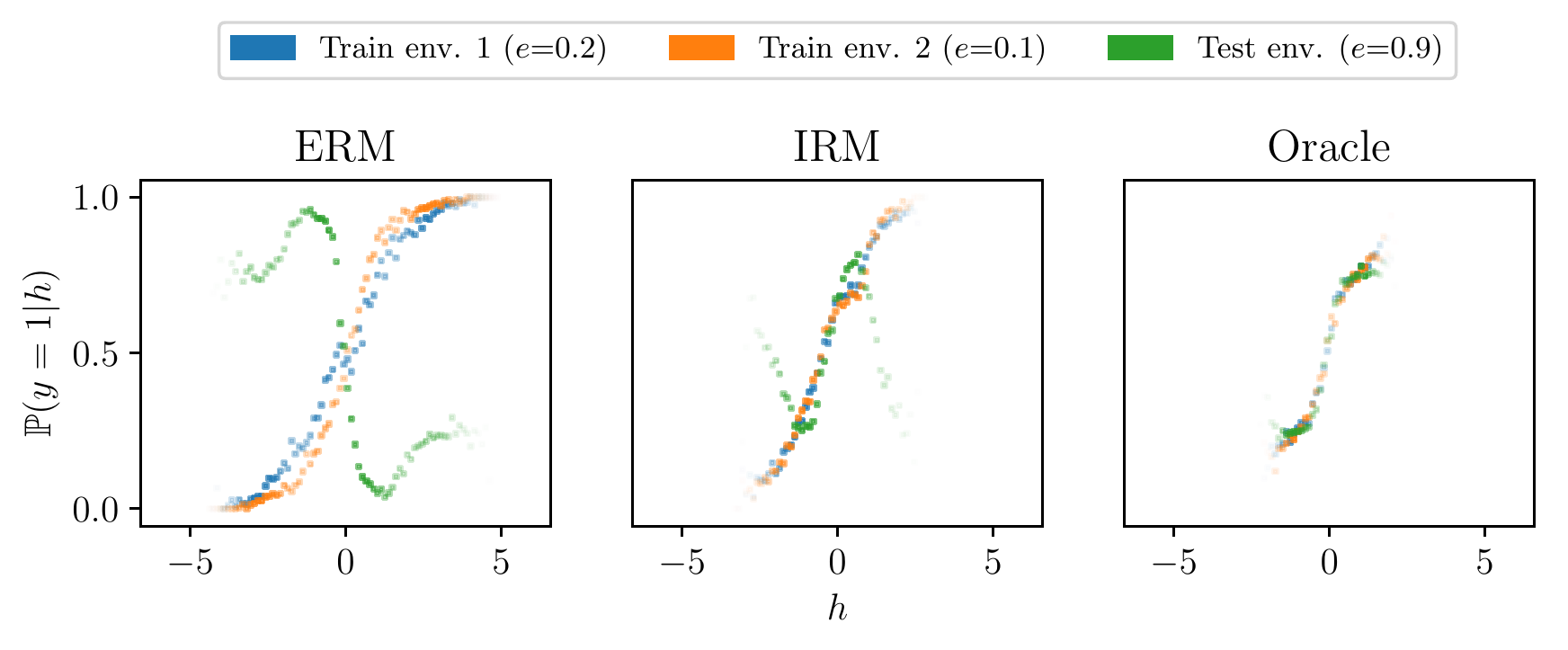}
  \caption{$P(y=1|h)$ as a function of $h$ for different models trained on Colored MNIST: (left) an ERM-trained model, (center) an IRM-trained model, and (right) an ERM-trained model which only sees grayscale images and therefore is perfectly invariant by construction. IRM learns approximate invariance from data alone and generalizes well to the test environment.}
  \label{fig:calibration}
\end{figure}

To better understand the behavior of these models, we take advantage of the fact that $h = \Phi(x)$ (the logit) is one-dimensional and $y$ is binary, and plot $P(y=1|h,e)$ as a function of $h$ for each environment and each model in \autoref{fig:calibration}.
We show each algorithm in a separate plot, and each environment in a separate color.
The figure shows that, whether considering only the two training environments or all three environments, the IRM model is closer to achieving invariance than the ERM model.
Notably, the IRM model does not achieve perfect invariance, particularly at the tails of the $P(h)$.
We suspect this is due to finite sample issues: given the small sample size at the tails, estimating (and hence minimizing) the small differences in $P(y|h, e)$ between training environments can be quite difficult, regardless of the method.

We note that conditional domain adaptation techniques which match $P(h|y,e)$ across environments could in principle solve this task equally well to IRM, which matches $P(y|h,e)$.
This is because the distribution of the causal features (the digit shapes) and $P(y|e)$ both happen to be identical across environments.
However, unlike IRM, conditional domain adaptation will fail if, for example, the distribution of the digits changes across environments.
We discuss this further in Appendix~\autoref{app:ada}.

Finally, \autoref{fig:calibration} shows that $P(y=1|h)$ cannot always be expressed with a linear classifier $w$.
Enforcing nonlinear invariances (Section~\ref{sub:nonlinw}) could prove useful here.

\section{Looking forward: a concluding dialogue}
\label{sec:outlook}
\direct{
    \refer{\ce} and \refer{\ci} are two graduate students studying the Invariant Risk Minimization (IRM) manuscript.
    Over a cup of coffee at a caf\'e in Palais-Royal, they discuss the advantages and caveats that invariance brings to Empirical Risk Minimization (ERM).
}

\begin{dialogue}
    \si I have observed that predictors trained with ERM sometimes absorb biases and spurious correlations from data. 
    This leads to undesirable behaviours when predicting about examples that do not follow the distribution of the training data. 

    \se I have observed that too, and I wonder what are the reasons behind such phenomena.
    After all, ERM is an optimal principle to learn predictors from empirical data!

    \si It is, indeed.
    But even when your hypothesis class allows you to find the empirical risk minimizer efficiently, there are some assumptions at play.
    First, ERM assumes that training and testing data are identically and independently distributed according to the same distribution.
    Second, generalization bounds require that the ratio between the capacity of our hypothesis class and the number of training examples $n$ tends to zero, as $n \to \infty$.
    Third, ERM achieves zero test error only in the realizable case ---that is, when there exists a function in our hypothesis class able to achieve zero error.
    I suspect that violating these assumptions leads ERM into absorbing spurious correlations, and that this is where invariance may prove useful.

    \se Interesting.
    Should we study the three possibilities in turn?

    \si Sure thing! But first, let's grab another cup of coffee.
\end{dialogue}

\direct{We also encourage the reader to grab a cup of coffee.}

\dialogueseparator

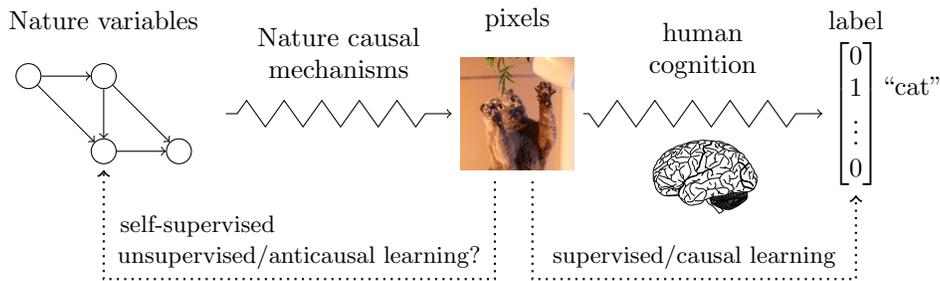
\begin{figure}[htpb]
    \centering
    \begin{tikzpicture}
        \node[] (a) at (1, 0.75)  {Nature variables}; 
        \node[draw=black, circle] (a) at (0, 0)  {}; 
        \node[draw=black, circle] (b) at (1, 0)  {}; 
        \node[draw=black, circle] (c) at (2, -1) {}; 
        \node[draw=black, circle] (d) at (1, -1) {}; 
        \node[] (anchor) at (2.5, -.5) {}; 
        \draw[->] (a) edge (b); 
        \draw[->] (a) edge (d); 
        \draw[->] (b) edge (c); 
        \draw[->] (d) edge (c); 
        \draw[->] (b) edge (d);

        \node[] (cat) at (6.5, -0.5)  {\includegraphics[width=1.5cm]{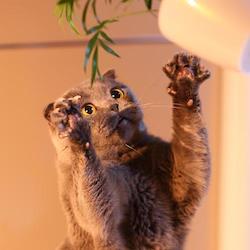}}; 
        \node[] (cattext) at (6.5, 0.75)  {pixels}; 
        
        \path [draw, <-, decorate, decoration={zigzag,amplitude=5pt,segment length=5mm,pre=lineto, pre length=10pt}] (cat) -- node [text width=2.5cm,midway,above,align=center,shift={(0,.35)}] {Nature causal mechanisms} (anchor);
        
        \node[] (labeltext) at (11, 0.75) {label};
        \node[] (label) at (11, -0.5)  {
        $\begin{bmatrix}
            0 \\
            1 \\
            \vdots \\
            0 
        \end{bmatrix}$};
        \node[] (label2) at (11.75, -0.13)  {``cat''};
        
        \path [draw, <-, decorate, decoration={zigzag,amplitude=5pt,segment length=5mm,pre=lineto, pre length=10pt}] (label) -- node [text width=2.5cm,midway,above,align=center,shift={(0,.35)}] {human cognition } (cat);

        \node[] (brain) at (9, -1.32) {\includegraphics[width=1.5cm]{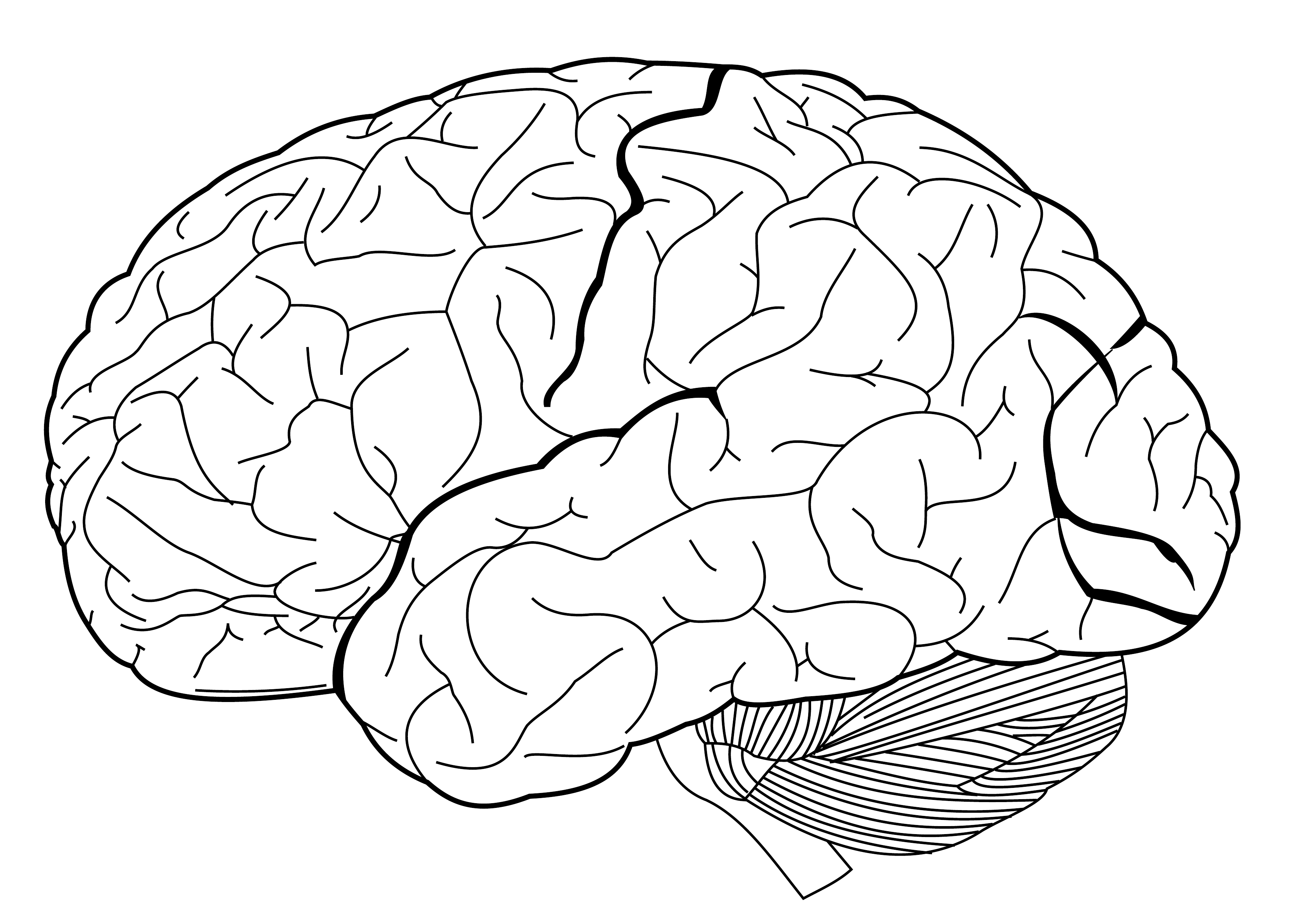}};

        \node [] at (8.85, -2.4) {\small supervised/causal learning};
        \path [draw, thick, dotted, ->] (6.7, -1.4) -- (6.7, -2.7) -- (11, -2.7) -- (label.south);
        
        \path [draw, thick, dotted, ->] (6.2, -1.4) -- (6.2, -2.7) -- (1, -2.7) -- (1, -1.25);
        \node [] at (3.6, -2.4) {\small unsupervised/anticausal learning?};
        \node [] at (2.25, -2) {\small self-supervised};

    \end{tikzpicture}
    \caption{All learning problems use empirical observations, here referred to as ``pixels''.
    Following a causal and cognitive process, humans produce labels.
    Therefore, supervised learning problems predicting annotations from observations are causal, and therefore $P(\text{label} \,|\, \text{pixel})$ is often invariant.
    Conversely, types of unsupervised and self-supervised learning trying to disentangle the underlying data causal factors of variation (Nature variables) should to some extent reverse the process generating observations (Nature mechanisms).
    This leads to anticausal learning problems, possibly with varying conditional distributions; an opportunity to leverage invariance.
    Cat picture by \url{www.flickr.com/photos/pustovit}.
    }
    \label{fig:anticausal}
\end{figure}

\begin{dialogue}
    \si First and foremost, we have the ``identically and independently distributed'' (iid) assumption.
    I once heard Professor Ghahramani refer to this assumption as ``the big lie in machine learning''.
    This is to say that all training and testing examples are drawn from the same distribution $P(X, Y) = P(Y | X) P(X)$.

    \se I see.
    This is obviously not the case when learning from multiple environments, as in IRM.
    Given this factorization, I guess two things are subject to change: either the marginal distribution $P(X)$ of my inputs, or the conditional distribution $P(Y|X)$ mapping those inputs into my targets. 

    \si That's correct. Let's focus first on the case where $P(X^e)$ changes across environments $e$.
    Some researchers from the field of domain adaptation call this \emph{covariate shift}.
    This situation is challenging when the supports of $P(X^e)$ are disjoint across environments.
    Actually, without a-priori knowledge, there is no reason to believe that our predictor will generalize outside the union of the supports of the training environments.

    \se A daunting challenge, indeed. How could invariance help here? 

    \si Two things come to mind.
    On the one hand, we could try to transform our inputs into some features $\Phi(X^e)$, as to match the support of all the training environments.
    Then, we could learn an invariant classifier $w(\Phi(X^e))$ on top of the transformed inputs.
    \direct{Appendix~D studies the shortcomings of this idea.}
    On the other hand, we could assume that the invariant predictor $w$ has a simple structure, that we can estimate given limited supports.
    The authors of IRM follow this route, by assuming linear classifiers on top of representations.

    \se I see!
    Even though the $P(X^e)$ may be disjoint, if there is a simple invariance satisfied for all training environments separately, it may also hold in unobserved regions of the space.
    I wonder if we could go further by assuming some sort of compositional structure in $w$, the linear assumption of IRM is just the simplest kind.
    I say this since compositional assumptions often enable learning in one part of the input space, and evaluating on another.

    \si It sounds reasonable!
    What about the case where $P(Y^e | X^e)$ changes?
    Does this happen in normal supervised learning?
    I remember attending a lecture by Professor Sch\"olkopf \citep{scholkopf2012causal, kilbertus2018generalization} where he mentioned that $P(Y^e | X^e)$ is often invariant across environments when $X^e$ is a cause of $Y^e$, and that it often varies when $X^e$ is an effect of $Y^e$.
    For instance, he explains that MNIST classification is anticausal: as in, the observed pixels are an effect of the mental concept that led the writer to draw the digit in the first place.
    IRM insists on this relation between invariance and causation, what do you think?

    \se I saw that lecture too.
    Contrary to Professor Sch\"olkopf, I believe that most supervised learning problems, such as image classification, are \emph{causal}.
    In these problems we predict human annotations $Y^e$ from pixels $X^e$, hoping that the machine imitates this cognitive process.
    Furthermore, the annotation process often involves multiple humans in the interest of making $P(Y^e | X^e)$ deterministic.
    If the annotation process is close to deterministic and shared across environments, predicting annotations is a causal problem, with an invariant conditional expectation.

    \si Oh!
    This means that in supervised learning problems about predicting annotations, $P(Y^e | X^e)$ is often stable across environments, so ERM has great chances of succeeding.
    This is good news: it explains why ERM is so good at supervised learning, and leaves less to worry about.
    \se However, if any of the other problems appear (disjoint $P(X^e)$, not enough data, not enough capacity), ERM could get in trouble, right?
    \si Indeed! Furthermore, in some supervised learning problems, the label is not necessarily created from the input.
    For instance, the input could be an X-ray image, and the target could be the result of a tumor biopsy on the same patient.
    Also, there are problems where we predict parts of the input from other parts of the input, like in self-supervised learning \cite{devlin2018bert}.
    In some other cases, we don't even have labels!
    This could include the unsupervised learning of the causal factors of variation behind $X^e$, which involves inverting the causal generative process of the data.
    In all of these cases, we could be dealing with anticausal problems, where the conditional distribution is subject to change across environments.
    Then, I expect searching for invariance may help by focusing on invariant predictors that generalize out-of-distribution.

    \se That is an interesting divide between supervised and unsupervised learning!
    \direct{Figure~\ref{fig:anticausal} illustrates the main elements of this discussion.}
\end{dialogue}

\dialogueseparator

\begin{dialogue}
    \se Secondly, what about the ratio between the capacity of our classifier and the number of training examples $n$?
    Neural networks often have a number of parameters on the same order of magnitude, or even greater, than the number of training examples \citep{zhang2016understanding}.
    In these cases, such ratio will not tend to zero as $n \to \infty$.
    So, ERM may be in trouble.

    \si That is correct.
    Neural networks are often over-parametrized, and over-parametrization carries subtle consequences.
    For instance, consider that we are using the pseudo-inverse to solve an over-parametrized linear least-squares problem, or using SGD to train an over-parametrized neural network.
    Amongst all the zero training error solutions, these procedures will prefer the solution with the smallest capacity \cite{wilson-marginal2017, benign}.
    Unfortunately, spurious correlations and biases are often simpler to detect than the true phenomenon of interest \cite{geirhos2018imagenet, texture1, texture2, texture3}.
    Therefore, low capacity solutions prefer exploiting those simple but spurious correlations.
    For instance, think about relying on large green textures to declare the presence of a cow on an image.

    \se The cows again!

    \si Always.
    Although I can give you a more concrete example. Consider predicting $Y^e$ from $X^e = (X^e_1, X^e_2)$, where:
    \begin{align*}
    Y^e   &\leftarrow 10^6 \cdot X^e_1 \alpha_1,\\
    X^e_2  &\leftarrow 10^6 \cdot Y^e \alpha_2^\top \cdot e,
    \end{align*}
    the coefficients satisfy $\| \alpha_1 \| = \| \alpha_2 \| = 1$, the training environments are $e = \{1, 10\}$, and we have $n$ samples for the $2n$-dimensional input $X$. 
    In this over-parametrized problem, the invariant regression from the cause $X_1$ requires large capacity, while the spurious regression from the effect $X_2$ requires low capacity.

    \se Oh! Then, the inductive bias of SGD would prefer to exploit the spurious correlation for prediction.
    In a nutshell, a deficit of training examples forces us into regularization, and regularization comes with the danger of absorbing easy spurious correlations. 
    But, methods based on invariance should realize that, after removing the nuisance variable $X_2$, the regression from $X_1$ is invariant, and thus interesting for out-of-distribution generalization.
    This means that invariance could sometimes help fight the issues of small data and over-parametrization.
    Neat!
\end{dialogue}

\dialogueseparator

\begin{dialogue}
    \si As a final obstacle to ERM, we have the case where the capacity of our hypothesis class is insufficient to solve the learning problem at hand.

    \se This sounds related to the previous point, in the sense that a model with low capacity will stick to spurious correlations, if these are easier to capture.

    \si That is correct, although I can see an additional problem arising from insufficient capacity.
    For instance, the only \emph{linear} invariant prediction rule to estimate the quadratic $Y^e = (X^e)^2$, where $X^e \sim \text{Gaussian}(0, e)$, is the null predictor $Y = 0 \cdot X$. 
    Even though $X$ is the only, causal, and invariance-eliciting covariate!

    \se Got it. Then, we should expect invariance to have a larger chance of success when allowing high capacity.
    For low-capacity problems, I would rely on cross-validation to lower the importance of the invariance penalty in IRM, and fall back to good old ERM.

    \si ERM is really withstanding the test of time, isn't it?

    \se Definitely.
    From what we have discussed before, I think ERM is specially useful in the realizable case, when there is a predictor in my hypothesis class achieving zero error.

    \si Why so?

    \se In the realizable case, the optimal invariant predictor has zero error across all environments.
    Therefore it makes sense, as an empirical principle, to look for zero training error across training environments.
    This possibly moves towards an optimal prediction rule on the union of the supports of the training environments.
    This means that achieving invariance across all environments using ERM is possible in the realizable case, although it would require data from lots of environments! 

    \si Wait a minute.
    Are you saying that achieving zero training error makes sense from an invariance perspective?

    \se In the realizable case, I would say so!
    Turns out all these people training neural networks to zero training error were onto something!
\end{dialogue}

\dialogueseparator

\noindent\direct{
    The barista approaches \refer{\ce} and \refer{\ci} to let them know that the caf\'e is closing.
  }

\begin{dialogue}
    \se Thank you for the interesting chat, \refer{\ci}. 

    \si The pleasure is mine! 

    \se One of my takeaways is that discarding spurious correlations is something doable even when we have access only to two environments.
    The remaining, invariant correlations sketch the core pieces of natural phenomena, which in turn form a simpler model.

    \si Simple models for a complex world. Why bother with the details, right?

    \se Hah, right.
    It seems like regularization is more interesting than we thought.
    IRM is a learning principle to discover \emph{unknown} invariances from data. This differs from typical regularization techniques to enforce \emph{known} invariances, often done by architectural choices (using convolutions to achieve translation invariance) and data augmentation.
    
    I wonder what other applications we can find for invariance.
    Perhaps we could think of reinforcement learning episodes as different environments, so we can learn robust policies that leverage the invariant part of behaviour leading to reward.

    \si That is an interesting one.
    I was also thinking that invariance has something to say about fairness.
    For instance, we could consider different groups as environments.
    Then, learning an invariant predictor means finding a representation such that the best way to treat individuals with similar relevant features is shared across groups.

    \se Interesting!
    I was also thinking that it may be possible to formalize IRM in terms of invariance and equivariance concepts from group theory. 
    Do you want to take a stab at these things tomorrow at the lab?

    \si Surely. See you tomorrow, \refer{\ce}.

    \se See you tomorrow!
\end{dialogue}

\noindent\direct{
    The students pay their bill, leave the caf\'e, and stroll down the streets of Paris, quiet and warm during the Summer evening.
}

\section*{Acknowledgements}
We are thankful to 
Francis Bach,
Marco Baroni,
Ishmael Belghazi,
Diane Bouchacourt,
Fran\c{c}ois Charton,
Yoshua Bengio,
Charles Blundell,
Joan Bruna,
Lars Buesing,
Soumith Chintala,
Kyunghyun Cho,
Jonathan Gordon,
Christina Heinze-Deml,
Ferenc Husz\'ar,
Alyosha Efros,
Luke Metz,
Cijo Jose,
Anna Klimovskaia,
Yann Ollivier,
Maxime Oquab,
Jonas Peters,
Alec Radford,
Cinjon Resnick,
Uri Shalit,
Pablo Sprechmann,
S\'onar festival,
Rachel Ward,
and
Will Whitney
for their help.

\bibliographystyle{plain}
\bibliography{outline}

\clearpage
\begin{appendices}

\section{Additional theorems}

\begin{theorem} \label{theo:genp1}
  Let $\Sigma_{X,X}^e := \EE_{X^e}[{X^e} {X^e}^\top] \in \SSS^{d \times d}_+$, with $\SSS^{d \times d}_+$ the space of symmetric positive semi-definite matrices, and $\Sigma_{X,\epsilon}^e := \EE_{X^e}[X^e \epsilon^e] \in \RR^d$.
  Then, for any arbitrary tuple $\left(\Sigma_{X, \epsilon}^e\right)_{e \in \Etrain} \in {\left(\RR^d\right)}^{|\Etrain|}$, the set 
  $$ \{(\Sigma_{X,X}^e)_{e \in \Etrain}\text{ such that $\Etrain$ does \emph{not} satisfy general position} \}$$
  has measure zero in $(\SSS_+^{d \times d})^{|\Etrain|}$.
\end{theorem}

\section{Proofs}
\label{app::proofs}

\subsection{Proof of Proposition~\ref{prop:robust}}

Let
\begin{align*}
f^\star &\in \min_{f} \max_{e \in \Etrain} R^e(f) - r_e,\\
M^\star     &= \max_{e \in \Etrain} R^e(f^\star) - r_e.
\end{align*}
Then, the pair $(f^\star, M^\star)$ solves the constrained optimization problem
\begin{align*}
\min_{f, M}& \quad M \nonumber \\
\text{s.t.}& \quad R^e(f) - r_e \leq M \quad \text{for all $e \in \Etrain$},
\end{align*}
with Lagrangian $L(f, M, \lambda) = M + \sum_{e \in \Etrain} \lambda^e (R^e(f) - r_e - M)$.
If the problem above satisfies the KKT differentiability and qualification conditions, then there exist $\lambda^e \geq 0$ with $\nabla_f L(f^\star, M^\star, \lambda) = 0$, such
that 
$$ \nabla_f|_{f=f^\star} \sum_{e \in \Etrain} \lambda^e R^e(f) = 0.$$

\subsection{Proof of Theorem~\ref{thm:solutions}}
\label{sub:proof_of_theorem_thm}

Let $\Phi\in\RR^{p\times d}$, $w \in \RR^{p}$, and $v = \Phi^\top w$.
The simultaneous optimization
\begin{equation}
\label{eq:pb-in-Sw}
    \forall e \quad w^\star \in \argmin_{w \in\R^{p}} R^e(w\circ\Phi)
\end{equation}
is equivalent to 
\begin{equation}
\label{eq:pb-in-v}
\forall e \quad v^\star \in \argmin_{v\in\mathcal{G}_{\Phi}} R^e(v),
\end{equation}
where $\mathcal{G}_{\Phi}=\{\Phi^\top w : w\in\R^{p}\}\subset\R^{d}$ is the
set of vectors~$v=\Phi^\top w$ reachable by picking any $w\in\R^{p}$.
It turns out that $\mathcal{G}_{\Phi}=\mathrm{Ker}(\Phi)^\perp$, that
is, the subspace orthogonal to the nullspace of $\Phi$. Indeed, for
all $v=\Phi^\top w\in\mathcal{G}_{\Phi}$ and all $x\in\mathrm{Ker}(\Phi)$,
we have $x^\top v=x^\top \Phi^\top w = (\Phi x)^\top w = 0$. Therefore
$\mathcal{G}_{\Phi}\subset\mathrm{Ker}(\Phi)^\perp$. Since both
subspaces have dimension $\mathrm{rank}(\Phi)=d-\mathrm{dim}(\mathrm{Ker}(\Phi))$,
they must be equal.

We now prove the theorem: let $v=\Phi^\top w$ where $\Phi\in\R^{p\times d}$
and~$w\in\R^{p}$ minimizes all~$R^e(w\circ\Phi)$. Since $v\in\mathcal{G}_\Phi$,
we have~$v\in\mathrm{Ker}(\Phi)^\perp$. Since~$w$ minimizes~$R^e(\Phi^\top {w})$, we can also write
\begin{equation}
\label{eq:vderivative}
\frac{\partial}{\partial w}\:R^e(\Phi^\top {w})
= \Phi\,\nabla{R^e}(\Phi^\top w) = \Phi \nabla{R^e}(v) =  0~.
\end{equation}
Therefore~$\nabla{R^e}(v)\in\mathrm{Ker}(\Phi)$.
Finally~$v^\top\nabla{R^e}(v)=w^\top \Phi\,\nabla{R^e}(\Phi^\top w)=0$.

Conversely, let $v\in\R^d$ satisfy~$v^\top\nabla{R^e}(v)=0$ for
all~$e\in\mathcal{E}$. Thanks to these orthogonality conditions, we can construct a
subspace that contains all the~$\nabla{R^e}(v)$ and is
orthogonal to~$v$. Let~$\Phi$ be any matrix whose nullspace satisfies
these conditions. Since $v\perp\mathrm{Ker}(\Phi)$, that is, $v\in\mathrm{Ker}(\Phi)^\perp=\mathcal{G}_{\Phi}$,
there is a vector $w \in \RR^{p}$ such that $v=\Phi^\top {w}$.
Finally, since $\nabla{R^e}(v)\in\mathrm{Ker}(\Phi)$, the
derivative~\eqref{eq:vderivative} is zero.

\subsection{Proof of Theorem~\ref{theo:lingen}}

Observing that $\Phi \, \EE_{X^e, Y^e}\left[{X^e} Y^e\right] = \Phi\, \EE_{X^e, \epsilon^e}[{X^e}({(\tilde S X^e)}^\top \gamma + \epsilon^e)]$, we re-write~\eqref{eq:theo_cond} as
\begin{equation}
  \Phi \left(\underbrace{\EE_{X^e}\left[{X^e} {X^e}^\top \right] (\Phi^\top w - \tilde S^\top \gamma) - \EE_{X^e, \epsilon^e}\left[{X^e} \epsilon^e\right]}_{:= q_e}\right) = 0.
    \label{eq:optmatchnew}
\end{equation}
To show that $\Phi$ leads to the desired invariant predictor $\Phi^\top w = \tilde S^\top \gamma$, we assume $\Phi^\top w \neq \tilde S^\top \gamma$ and reach a contradiction.
First, by Assumption~\ref{a:gpos}, we have $\text{dim}(\text{span}(\{q_e\}_{e \in \Etrain})) > d - r$.
Second, by~\eqref{eq:optmatchnew}, each $q_e \in \text{Ker}(\Phi)$.
    Therefore, it would follow that $\text{dim}(\text{Ker}(\Phi)) > d-r$, which contradicts the assumption that $\text{rank}(\Phi) = r$.

\subsection{Proof of Theorem~\ref{theo:genp1}}
  Let $m = |\Etrain|$, and define $G: \RR^d \setminus \{0\} \rightarrow \RR^{m \times d}$ as
  $ \left(G(x)\right)_{e, i} = \left(\Sigma^e_{X,X} x - \Sigma_{X,\epsilon}^e\right)_i.$

  Let $W = G\left(\RR^d \setminus \{0\} \right) \subseteq \RR^{m \times d}$, which is a linear manifold of dimension at most $d$, missing a single point (since $G$ is affine, and its input has dimension $d$).

  For the rest of the proof, let $(\Sigma^e_{X, \epsilon})_{e \in \Etrain} \in {\RR^{d}}^{|\Etrain|}$ be arbitrary and fixed.  
  We want to show that for generic $(\Sigma^e_{X,X})_{e \in \Etrain}$, if $m > \frac{d}{r} + d - r$, the matrices $G(x)$ have
  rank larger than $d - r$. Analogously, if $\text{LR}(m, d, k) \subseteq \RR^{m \times d}$ is the set of $m \times d$ matrices with rank $k$, we want to show
  that $W \cap \text{LR}(m, d, k) = \emptyset$ for all $k < d - r$.

  We need to prove two statements.
  First, that for generic$(\Sigma^e_{X,X})_{e \in \Etrain}$
  $W$ and $\text{LR}(m, d, k)$ intersect transversally as manifolds, or don't intersect at all. This will be a standard argument using Thom's transversality theorem.
  Second, by dimension counting, that if $W$ and $\text{LR}(m, d, k)$ intersect transversally, and $k < d - r$, $m > \frac{d}{r} + d - r$, then
  the dimension of the intersection is negative, which is a contradiction and thus $W$ and $\text{LR}(m, d, k)$ cannot intersect.

  We then claim that $W$ and $\text{LR}(m, d, k)$ are transversal for generic $(\Sigma^e_{X,X})_{e \in \Etrain}$.
    To do so, define $$F: (\RR^d \setminus \{0\}) \times \left(\SSS_+^{d \times d}\right)^m \rightarrow \RR^{m \times d},$$
  $$F\left(x, \left(\Sigma^e_{X,X}\right)_{e \in \Etrain}\right)^{e'}_l = \left(\Sigma_{X,X}^{e'} x - \Sigma_{X,\epsilon}^{e'}\right)_l$$
  If we show that $\nabla_{x, \Sigma_{X,X}} F: \RR^d \times (\SSS^{d \times d})^m \rightarrow \RR^{m \times d}$ is a surjective linear transformation,
  then $F$ is transversal to any submanifold of $\RR^{m \times d}$ (and in particular
  to $\text{LR}(m, d, k)$). By the Thom transversality theorem, this implies that the set of $\left(\Sigma^e_{X,X}\right)_{e \in \Etrain}$ such that $W$ is not transversal
  to $\text{LR}(m, d, k$) has measure zero in $\SSS_{+}^{d \times d}$, proving our first statement.

  Next, we show that $\nabla_{x, \Sigma_{X,X}} F$ is surjective. This follows by 
  by showing that $\nabla_{\Sigma_{X,X}} F: (\SSS^{d \times d})^m  \rightarrow \RR^{m \times d}$
  is surjective, since adding more columns to this matrix can only increase its rank.
  We then want to show that the linear map $\nabla_{\Sigma_{X,X}} F: (\SSS^{d \times d})^m  \rightarrow \RR^{m \times d}$ is surjective.
  To this end, we can write:
  $ \partial_{\Sigma_{i,j}^e} F^{e'}_l = \delta_{e,e'}\left(\delta_{l,i} x_j + \delta_{l,j} x_i\right),$
  and let $C \in \RR^{m \times d}$. We want to construct a $D \in \left(\SSS^{d \times d}\right)^m$ such that
  $$ C^{e'}_l = \sum_{i, j, e} \delta_{e,e'} \left(\delta_{l,i} x_j + \delta_{l,j} x_i\right) D^e_{i,j}. $$
  The right hand side equals
  \begin{align*}
    \sum_{i, j, e} \delta_{e,e'} \left(\delta_{l,i} x_j + \delta_{l,j} x_i\right) D^e_{i,j} &= \sum_{j} D^{e'}_{l,j} x_j +\sum_{i} D^{e'}_{i,l} x_i 
      = (D^{e'} x)_l + (x D^{e'})_l
  \end{align*}
  If $D^{e'}$ is symmetric, this equals $(2 D^e x)_l$. Therefore, we only need to show that for any vector $C^e \in \RR^d$, there is a symmetric matrix
  $D^e \in \SSS^{d \times d}$ with $C^e = D^e x$.
  To see this, let $O \in \RR^{d \times d}$ be an orthogonal transformation such that $O x$ has no zero entries,
  and name $v = O x, w^e = O C^e$. Furthermore, let $E^e \in \RR^{d \times d}$ be the diagonal matrix with entries $E^e_{i,i} = \frac{w^e_i}{v_i}$. Then,
  $C^e = O^T E^e O x$. By the spectral theorem, $O^T E^e O$ is symmetric, showing that $\nabla_{\Sigma_{X,X}} F: (\SSS^{d \times d})^m  \rightarrow \RR^{m \times d}$
  is surjective, and thus that $W$ and $\text{LR}(m, d, k)$ are transversal for almost any $\left(\Sigma_{X,X}^e\right)_{e \in \Etrain}$.

  By transversality, we know that $W$ cannot intersect $\text{LR}(m, d, k)$ if $\dim(W) + \dim\left(\text{LR}(m, d, k)\right) - \dim\left(\RR^{m \times d}\right) < 0$.
  By a dimensional argument (see \cite{lee2003introduction}, example 5.30), it follows that 
  $\codim(\text{LR}(m, d, k)) = \dim\left(\RR^{m \times d}\right) - \dim\left(\text{LR}(m, d, k)\right) = (m - k) (d - k)$.
  Therefore, if $k < d - r$ and $m > \frac{d}{r} + d - r$, it follows that
  \begin{align*}
      \dim(W) + \dim\left(\text{LR}(m, d, k)\right) - \dim\left(\RR^{m \times d}\right) &= \dim(W) - \codim\left(\text{LR}(m, d, k)\right) \\
      &\leq d - (m - k)(d - k) \\
      &\leq d - (m - (d - r))(d - (d - r)) \\
      &= d - r (m - d + r) \\
      &< d - r\left(\left(\frac{d}{r} + d - r\right)-d + r\right) \\
      &= d - d = 0.
  \end{align*}
  Therefore, $W \cap \text{LR}(m, d, k) = \emptyset$ under these conditions, finishing the proof.

\section{Failure cases for Domain Adaptation}
\label{app:ada}

Domain adaptation \cite{bendavid-domain} considers labeled data from a source environment $e_s$ and unlabeled data from a target environment $e_t$ with the goal of training a classifier that works well on $e_t$.
Many domain adaptation techniques, including the popular Adversarial Domain Adaptation \citep[ADA]{ganin2016domain}, proceed by learning a feature representation $\Phi$ such that (i) the input marginals $P(\Phi(X^{e_s})) = P(\Phi(X^{e_t}))$, and (ii) the classifier $w$ on top of $\Phi$ predicts well the labeled data from $e_s$.
Thus, are domain adaptation techniques applicable to finding invariances across multiple environments?

One shall proceed cautiously, as there are important caveats.
For instance, consider a binary classification problem, where the only difference between environments is that $P(Y^{e_s} = 1) = \frac{1}{2}$, but $P(Y^{e_t} = 1) = \frac{9}{10}$.
Using these data and the domain adaptation recipe outlined above, we build a classifier $w \circ \Phi$.
Since domain adaptation enforces $P(\Phi(X^{e_s})) = P(\Phi(X^{e_t}))$, it consequently enforces $P(\hat{Y}^{e_s}) = P(\hat{Y}^{e_t})$, where $\hat{Y}^e = w(\Phi(X^e))$, for all $e \in \{ e_s, e_t \}$.
Then, the classification accuracy will be at most 20\pc{}.
This is worse than random guessing, in a problem where simply training on the source domain leads to a classifier that generalizes to the target domain.

Following on this example, we could think of applying conditional domain adaptation techniques \cite[C-ADA]{Li_2018_ECCV}.
These enforce one invariance $P(\Phi(X^{e_s}) | Y^{e_s}) = P(\Phi(X^{e_t}) | Y^{e_t})$ per value of $Y^e$.
Using Bayes rule, it follows that C-ADA enforces a stronger condition than invariant prediction when $P(Y^{e_s}) = P(Y^{e_t})$.
However, there are general problems where the invariant predictor cannot be identified by C-ADA.

To see this, consider a discrete input feature $X^e \sim P(X^e)$, and a binary target $Y^e = F(X^e) \oplus \text{Bernoulli}(p)$.
This model represents a generic binary classification problem with label noise.
Since the distribution $P(X^e)$ is the only moving part across environments, the trivial representation $\Phi(x) = x$ elicits an invariant prediction rule.
Assuming that the discrete variable $X^e$ takes $n$ values, we can summarize $P(X^e)$ as the probability $n$-vector $p^{x, e}$.
Then, $\Phi(X^e)$ is also discrete, and we can summarize its distribution as the probability vector $p^{\phi, e} = A_\phi p^{x, e}$, where $A_\phi$ is a matrix of zeros and ones.
By Bayes rule,
\begin{align*}
    \pi^{\phi, e} := P(\Phi(X^e) | Y^e=1) = \frac{P(Y^e = 1 | \Phi(X^e)) \odot p^{\phi, e}}{\langle P(Y^e = 1 | \Phi(X^e)), p^{\phi, e} \rangle} = \frac{\left( A_\Phi \left(v \odot p^{x, e} \right)\right)\odot (A_\Phi p^{x,e})}{\langle\left( A_\Phi \left(v \odot p_{x, e} \right)\right), A_\Phi p^{x, e} \rangle},
\end{align*}
where $\odot$ is the entry-wise multiplication, $\langle , \rangle$ is the dot product, and $v := P(Y^e = 1 | X^e)$ does not depend on $e$.
Unfortunately for C-ADA, it can be shown that the set $\Pi_\phi := \{ (p^{x, e}, p^{x, e'}) : \pi^{\phi, e} = \pi^{\phi, e'} \}$ has measure zero.
Since the union of sets with zero measure has zero measure, and there exists only a finite amount of possible $A_\phi$, the set $\Pi_\phi$ has measure zero for \emph{any} $\Phi$. 
In conclusion and almost surely, C-ADA disregards any non-zero data representation eliciting an invariant prediction rule, regardless of the fact that the trivial representation $\Phi(x) = x$ achieves such goal.

As a general remark, domain adaptation is often justified using the bound \cite{bendavid-domain}:
\begin{equation*}
    \text{Error}^{e_t}(w \circ \Phi) \leq \text{Error}^{e_s}(w \circ \Phi) + \text{Distance}(\Phi(X^{e_s}), \Phi(X^{e_t})) + \lambda^\star.
\end{equation*}
Here, $\lambda^\star$ is the error of the optimal classifier in our hypothesis class, operating on top of $\Phi$, summed over the two domains.
Crucially, $\lambda^\star$ is often disregarded as a constant, justifying the DA goals (i, ii) outlined above.
But, $\lambda^\star$ depends on the data representation $\Phi$, instantiating a third trade-off that it is often ignored.
For a more in depth analysis of this issue, we recommend \citep{Johansson2019SupportAI}.

\section{Minimal implementation of IRM in PyTorch}
\label{sec:code}
\lstset{
    language=Python,
    basicstyle=\footnotesize\ttfamily,
    keywordstyle=\color{blue}\ttfamily,
    stringstyle=\color{red}\ttfamily,
    commentstyle=\color{green}\ttfamily,
    morecomment=[l][\color{magenta}]{\#}
}

\begin{lstlisting}
import torch
from torch.autograd import grad

def compute_penalty(losses, dummy_w):
    g1 = grad(losses[0::2].mean(), dummy_w, create_graph=True)[0]
    g2 = grad(losses[1::2].mean(), dummy_w, create_graph=True)[0]
    return (g1 * g2).sum()

def example_1(n=10000, d=2, env=1):
    x = torch.randn(n, d) * env
    y = x + torch.randn(n, d) * env
    z = y + torch.randn(n, d)
    return torch.cat((x, z), 1), y.sum(1, keepdim=True)

phi = torch.nn.Parameter(torch.ones(4, 1))
dummy_w = torch.nn.Parameter(torch.Tensor([1.0]))

opt = torch.optim.SGD([phi], lr=1e-3)
mse = torch.nn.MSELoss(reduction="none")

environments = [example_1(env=0.1),
                example_1(env=1.0)]

for iteration in range(50000):
    error = 0
    penalty = 0
    for x_e, y_e in environments:
        p = torch.randperm(len(x_e))
        error_e = mse(x_e[p] @ phi * dummy_w, y_e[p])
        penalty += compute_penalty(error_e, dummy_w)
        error += error_e.mean()

    opt.zero_grad()
    (1e-5 * error + penalty).backward()
    opt.step()

    if iteration % 1000 == 0:
        print(phi)
\end{lstlisting}
\end{appendices}

\end{document}